\documentclass[11pt]{article}
\usepackage[preprint]{acl}
\usepackage{times}
\usepackage{latexsym}
\usepackage[T1]{fontenc}
\usepackage[utf8]{inputenc}
\usepackage{microtype}
\usepackage{inconsolata}

\usepackage{amsmath,amsfonts,bm}

\def\eqref#1{equation~\ref{#1}}

\def\1{\bm{1}}

\DeclareMathAlphabet{\mathsfit}{\encodingdefault}{\sfdefault}{m}{sl}
\SetMathAlphabet{\mathsfit}{bold}{\encodingdefault}{\sfdefault}{bx}{n}

\usepackage{amssymb}
\usepackage{booktabs,multirow,adjustbox,graphicx,makecell,pifont}
\usepackage[table]{xcolor}
\usepackage[most]{tcolorbox}
\usepackage{placeins}
\usepackage{algorithm}
\usepackage{algpseudocode}
\usepackage{enumitem}
\usepackage{subcaption}

\usepackage{hyperref}
\usepackage{fontawesome5}

\usepackage{longtable}
\usepackage{fvextra}
\usepackage{CJKutf8}
\definecolor{okgreen}{HTML}{2E7D32}
\definecolor{warnamber}{HTML}{B26A00}
\definecolor{failred}{HTML}{B3261E}

\newtcolorbox{tracebox}[3][]{%
  enhanced, sharp corners, boxrule=0.7pt, width=\linewidth,
  left=4pt, right=4pt, top=3pt, bottom=3pt,
  colframe=#2, colback=#2!3, coltitle=white, colbacktitle=#2,
  fonttitle=\footnotesize\bfseries, title={#3}, #1}

\newtcolorbox{problembox}[1]{%
  enhanced, sharp corners, boxrule=0.7pt, width=\textwidth,
  left=6pt, right=6pt, top=4pt, bottom=4pt,
  colframe=black!70, colback=black!3, coltitle=white, colbacktitle=black!70,
  fonttitle=\footnotesize\bfseries, title={#1}}

\newcommand{\tracecell}[1]{%
  \begin{minipage}[t]{0.315\textwidth}\raggedright #1\end{minipage}}

\newcommand{\tracegap}{[...]}
\newcommand{\tracestage}[1]{\par #1:\ }
\newcommand{\traceobs}[1]{%
  \par\smallskip
  \hrule
  \smallskip
  \noindent \textit{Observation:} #1%
}
\newcommand{\tracemethod}[3]{#1\par #2 tokens\par #3}
\newcommand{\tracesymbol}[1]{\ensuremath{\boxed{?}}}

\title{Unlocking Structured Pruning for Reasoning Language Models via Attention-Preserving MLP Trimming}

\author{
Woosang Lim\textsuperscript{1,*},
Haneul Yoo\textsuperscript{2,*},
Vincent-Daniel Yun\textsuperscript{3,*},
Sungjoo Yoo\textsuperscript{1}\thanks{%
    \begin{tabular}[t]{@{}l@{}}
        Corresponding author: sungjoo.yoo@gmail.com. \\
        \textsuperscript{*}These authors contributed equally.
    \end{tabular}%
}
\\ \\
\textsuperscript{1}Seoul National University, Republic of Korea \\
\textsuperscript{2}New York University, United States \\
\textsuperscript{3}University of Southern California, United States \\
\\ 
\href{https://github.com/daniel-eai/LayerTrim}
{\faGithub\ GitHub}
}

\vspace{0.5cm}

\begin{document}
\setcounter{footnote}{1}

\maketitle

\begin{abstract}
Structured pruning reduces LLM inference costs by removing model components to produce smaller dense models. However, existing methods can cause reasoning performance to collapse on long-form reasoning tasks such as AIME. We propose \textit{LayerTrim}, a training-free structured pruning method that identifies redundant decoder layers and selectively trims their MLP features while preserving attention and decoder depth. LayerTrim is the first training-free structured pruning method designed to preserve long-form reasoning without gradients or recovery training. Across three reasoning language models and five benchmarks, LayerTrim outperforms existing structured pruning methods in most evaluated settings, improving average accuracy by up to 21.4 percentage points over RESP. On \texttt{Qwen3-14B}, LayerTrim achieves 1.24$\times$ faster prefill and 19.7\% lower peak GPU memory than the dense model, while better preserving reasoning ability. It also achieves the shortest average generation length among evaluated pruned models, while retaining the highest average accuracy. These results demonstrate that LayerTrim improves reasoning efficiency through both structural compression and reduced token generation.

\end{abstract}

\section{Introduction}

\label{sec:introduction}
\label{sec:intro}

Reasoning language models (RLMs) achieve strong performance on complex tasks
through chain-of-thought (CoT) reasoning~\citep{deepseekai2025deepseekr1,yang2025qwen3}.
However, generating long reasoning sequences incurs high inference
costs, motivating more efficient inference.
Structured pruning reduces LLM inference costs by removing
model components to produce smaller dense models~\citep{ma2023llmpruner,an2024flap}.
However, applying pruning methods developed for non-reasoning LLMs
to RLMs can cause performance to collapse
on tasks requiring long chains of thought, such as
AIME~\citep{ding2026rethinking}.

\begin{figure}[t]
\centering
\includegraphics[width=\linewidth]{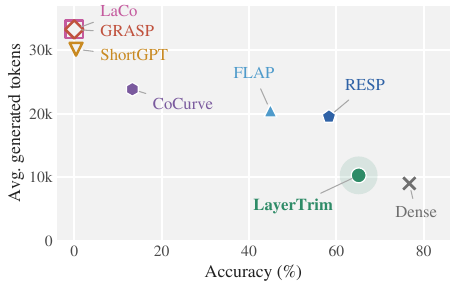}
\caption{Accuracy versus generated tokens per response
on \texttt{Qwen3-14B} with a pruning budget equivalent to 9 of its
40 decoder layers. Both metrics are averaged across five
benchmarks, with a maximum generation length of 32K tokens.}
\label{fig:hero_fig_intro}
\end{figure}

\begin{figure*}[t]
    \centering
    \includegraphics[width=\textwidth]{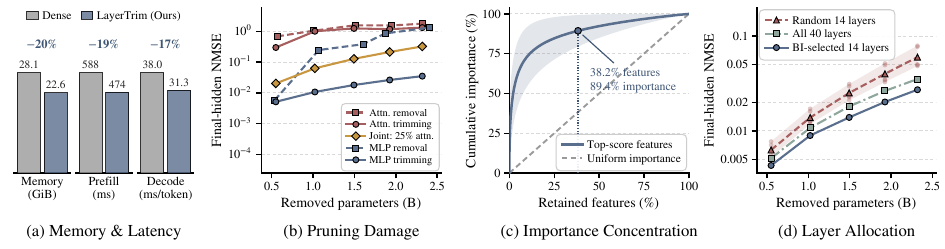}
    \caption{Observations motivating LayerTrim on \texttt{Qwen3-14B}.
    (a) LayerTrim reduces memory usage and inference latency
    with a pruning budget equivalent to 9 of 40 decoder layers.
    (b) MLP trimming causes less representation error than
    attention pruning, module removal, or joint pruning.
    (c) MLP feature importance is concentrated in a subset of features.
    (d) Block Influence (BI)~\citep{men2024shortgpt}-based layer
    selection reduces error compared with random or all-layer pruning.
    Diagnostic settings are given in Appendix~\ref{app:attention-surrogate-analysis}.}
    \label{fig:layertrim_observations}
\end{figure*}

Figure~\ref{fig:hero_fig_intro} illustrates this failure,
where existing structured pruning methods lead to longer
reasoning traces and lower accuracy.
To investigate this degradation, we analyze pruning strategies
on \texttt{Qwen3-14B}~\citep{yang2025qwen3}.
Our analysis reveals three observations
(Figure~\ref{fig:layertrim_observations}).
First, MLP trimming causes less representation error than
attention pruning or entire module removal.
Second, MLP feature importance is concentrated in a subset of features.
Third, concentrating MLP trimming in redundant layers produces
less error than distributing the same budget across all layers.
These findings motivate selective MLP compression while
preserving attention and decoder depth.

Based on these findings, we propose \textbf{LayerTrim},
a training-free structured pruning method that identifies redundant
decoder \textbf{layers} and selectively \textbf{trims} their MLP features
while preserving attention and decoder depth.
LayerTrim is the first training-free
structured pruning method designed to preserve long-form reasoning
and code generation performance without recovery training.
It uses dense-model reasoning traces
\citep{lucas2025rac,wang2025resp} to select layers using
Block Influence~\citep{men2024shortgpt} and retain MLP features
with the largest expected squared output contributions.

Across three RLMs and five benchmarks,
LayerTrim outperforms existing structured pruning methods in
most evaluated settings, improving average accuracy by up to
21.4 percentage points over RESP~\citep{wang2025resp} under matched parameter budgets.
On \texttt{Qwen3-14B}, LayerTrim achieves higher accuracy while generating
fewer tokens on average than the other evaluated pruning methods
(Figure~\ref{fig:hero_fig_intro}).
It also reduces peak GPU memory, prefill latency, and decode latency
by approximately 20\%, 19\%, and 17\%, respectively, relative to the
dense model at a pruning budget equivalent to 9 of 40 decoder layers
(Figure~\ref{fig:layertrim_observations}(a)).

Our contributions are as follows:
\begin{itemize}[leftmargin=*]
    \item We identify attention preservation and selective MLP
    trimming as strategies for reducing representation error
    compared with attention pruning or entire module removal.

    \item We introduce \textit{LayerTrim}, a training-free
    structured pruning method for long-form reasoning without gradients or recovery training.

    \item We demonstrate improved reasoning efficiency,
    achieving higher accuracy with fewer generated tokens
    and better
    preserving reasoning completion.
\end{itemize}

\section{Related Work}

\label{sec:related_work}

\paragraph{Structured pruning of LLMs.}
Unstructured pruning methods such as SparseGPT~\citep{frantar2023sparsegpt}
and Wanda~\citep{sun2023wanda} remove individual weights,
often requiring sparse kernels for practical acceleration.
In contrast, structured pruning removes computational units
to produce smaller dense models~\citep{depthpruning}.
LLM-Pruner~\citep{ma2023llmpruner} and HFPrune~\citep{zhu2026hfprune}
use gradient- or Taylor-based importance criteria,
while FLAP~\citep{an2024flap} compensates for pruning errors.
Sheared LLaMA~\citep{xia2024sheared} incorporates additional training,
and CoCurve~\citep{gong2026cocurve} considers interactions between
structured units. Depth pruning methods such as ShortGPT~\citep{men2024shortgpt}
identify redundant decoder layers using Block Influence,
while LaCo~\citep{yang2024laco} and GRASP~\citep{liu2025grasp}
reduce model depth through layer merging or replacement.
LoRP~\citep{lorp} exploits inter-layer redundancy and
architectural locality to identify layers for training-free
depth pruning. However, these methods do not explicitly consider long-form CoT reasoning,
and removing entire layers can degrade reasoning performance.

\paragraph{Pruning RLMs.}
Pruning methods developed for non-reasoning LLMs can degrade
reasoning performance~\citep{ding2026rethinking}.
RAC~\citep{lucas2025rac} uses on-policy reasoning traces for
activation reconstruction, while OBC-Prune~\citep{obc-prune-rac2}
extends reasoning-aware calibration through outcome-based
token weighting. Both rely on unstructured pruning,
requiring sparse kernel support for practical acceleration.
RESP~\citep{wang2025resp} uses self-generated reasoning traces,
gradient-based importance, and progressive recalibration
for structured pruning, but retains lower reasoning accuracy
than dense models across the evaluated benchmarks.

Preserving long-form reasoning remains a challenge for existing structured pruning approaches. LayerTrim addresses this challenge by selectively trimming MLP features while preserving attention and decoder depth.

\section{Method}
\subsection{Motivation}

\label{sec:pruning_observations}
We first investigate three structural pruning decisions for RLMs: (1) which components to prune, (2) which MLP features to retain, and (3) how to allocate pruning across layers. Using \texttt{Qwen3-14B} and AIME reasoning traces generated by the dense model, we characterize MLP feature contributions and compare representation errors across pruning settings.

\paragraph{Pruning targets.}
Methods such as RESP prune attention and MLPs jointly. We separately compare attention and MLP removal, attention and MLP trimming, and joint trimming of both. Removal deletes modules selected by low Block Influence, whereas trimming reduces internal width across all 40 decoder layers.

Across all tested budgets, MLP-only trimming produces lower final-hidden representation error than module removal, attention trimming, or joint trimming of attention and MLPs (Figure~\ref{fig:layertrim_observations}(b)). These results motivate preserving attention and decoder depth while reducing MLP width.

\paragraph{Importance concentration within MLPs.}
To select features for width reduction, we express the MLP output as a sum of intermediate feature terms. For a SwiGLU MLP~\citep{shazeer2020glu} with intermediate width $I$, let $\phi^{(\ell)}(x)=\operatorname{SiLU}(W_{\mathrm{gate}}^{(\ell)}x) \odot(W_{\mathrm{up}}^{(\ell)}x)\in\mathbb{R}^{I}$ denote its intermediate activation vector. The $j$th feature contributes $c_j^{(\ell)}(x)=\phi_j^{(\ell)}(x)W_{\mathrm{down}}^{(\ell)}[:,j]$ to the MLP output, yielding $\mathrm{MLP}^{(\ell)}(x)=\sum_{j=1}^{I}c_j^{(\ell)}(x)$. 

We define the feature importance score as the expected squared norm of $c_j^{(\ell)}(x)$:
\begin{equation}
    \begin{aligned}
    s_j^{(\ell)}
    &= \mathbb{E}_{\mathcal D}\!\left[
        \|c_j^{(\ell)}(x)\|_2^2
      \right] \\
    &= \mathbb{E}_{\mathcal D}\!\left[
        (\phi_j^{(\ell)}(x))^2
      \right]
      \|W_{\mathrm{down}}^{(\ell)}[:,j]\|_2^2
    \end{aligned}
    \label{eq:layertrim_score}
\end{equation}
Here, $\mathcal D$ denotes calibration token positions, and $x$ is the dense-model MLP input. Figure~\ref{fig:layertrim_observations}(c) shows that the top 38.2\% of features account for 89.4\% of the total feature importance, supporting their retention during MLP width reduction.

\paragraph{Redundancy-aware budget allocation.} Feature importance determines which dimensions to retain within an MLP, while pruning budget allocation determines which layers undergo width reduction. Figure~\ref{fig:layertrim_observations}(d) compares trimming all 40 MLPs, 14 randomly selected MLPs, and 14 MLPs selected by low Block Influence under matched parameter budgets. Trimming the MLPs selected by low Block Influence produces lower final-hidden representation error than either alternative. The comparison with random selection indicates that layer selection matters beyond the number of trimmed MLPs.

\subsection{LayerTrim}
\label{sec:method}
\label{sec:layertrim_procedure}

\begin{figure*}[t]
    \centering
    \includegraphics[width=\textwidth]{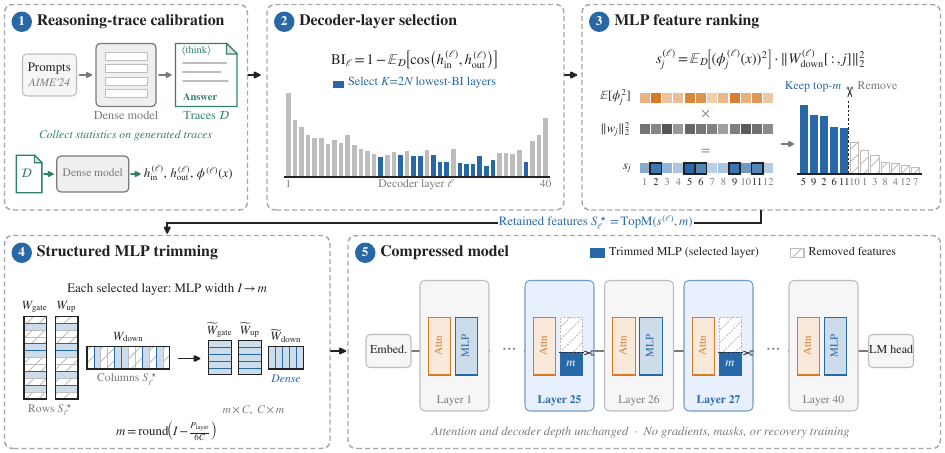}
    \caption{Overview of LayerTrim. LayerTrim collects layer-level
    and MLP feature-level statistics from reasoning traces
    generated by the dense model. It selects redundant decoder
    layers and retains MLP features with larger output contributions.
    The selected MLPs are reduced to smaller dense layers
    while preserving attention and decoder depth.}
    \label{fig:place}
\end{figure*}

Guided by these observations,
LayerTrim reduces MLP width while preserving attention and decoder depth.
It retains high-importance features and concentrates pruning in less
influential layers.
Figure~\ref{fig:place} summarizes five stages: reasoning-trace calibration,
decoder-layer selection, MLP feature ranking, structured MLP trimming,
and the resulting compressed model.
Algorithm~\ref{alg:layertrim} in Appendix~\ref{app:layertrim-algorithm}
gives the complete procedure.

\paragraph{Reasoning-trace calibration.} Following prior work on pruning RLMs~\citep{lucas2025rac,wang2025resp},
we collect statistics through teacher-forced forward passes over
reasoning traces generated by the dense model.
For each layer, we record decoder-block inputs and outputs and MLP
intermediate activations. These statistics support layer selection
and feature ranking without gradients.

\paragraph{Decoder-layer selection.}
To concentrate pruning in less influential layers, we use \textit{Block Influence}
(BI)~\citep{men2024shortgpt}:
\begin{equation}
    \mathrm{BI}_{\ell}
    =1-\mathbb{E}_{\mathcal D}\!\left[
      \cos\!\left(h_{\mathrm{in}}^{(\ell)},
                  h_{\mathrm{out}}^{(\ell)}\right)\right]
    \label{eq:layertrim_bi}
\end{equation}
where $h_{\mathrm{in}}^{(\ell)}$ and $h_{\mathrm{out}}^{(\ell)}$
are decoder block inputs and outputs.
Low BI indicates a small directional change across a block.
For a budget of $N$ layer-equivalents, we select the $K=2N$ lowest-BI
layers, denoted $\mathcal L_K$,
and modify only their MLPs.
This choice spreads the pruning budget across more MLPs rather than
applying stronger width reduction to fewer layers.

\paragraph{MLP feature ranking.}
Let $m$ denote the number of features retained in each selected MLP.
Let $C$ and $I$ be the hidden and MLP intermediate widths, and
$P_{\mathrm{layer}}$ the prunable parameter count of one decoder layer.
Each removed feature eliminates $C$ weights from each of the gate, up, and down projections, totaling $3C$ parameters per feature and $3CK(I-m)$ across the $K$ selected MLPs.
Setting this count equal to the budget, $3CK(I-m) = N P_{\mathrm{layer}}$, and using $K = 2N$ gives
\begin{equation}
    m=\operatorname{round}\!\left(
        I-\frac{P_{\mathrm{layer}}}{6C}
    \right)
    \label{eq:keep_width}
\end{equation}

For each selected MLP, we compute feature importance $s_j^{(\ell)}$
using Equation~\ref{eq:layertrim_score}.
We rank features by this score and retain the top $m$,
$S_\ell^\star=\operatorname{TopM}(s^{(\ell)},m)$.
To reduce calibration overhead, we approximate the squared MLP output
change using the summed scores of removed features.

\paragraph{Structured MLP trimming.}
We physically remove the remaining $I-m$ features from each selected MLP.
For each $\ell\in\mathcal L_K$, we retain rows $S_\ell^\star$
of the gate and up projections and columns $S_\ell^\star$
of the down projection.
The reduced projections have shapes $m\times C$ and $C\times m$.
The retained weights form a compressed checkpoint with smaller dense MLPs.
Attention modules, unselected MLPs, and decoder depth remain unchanged.
No inference-time masks, fine-tuning, distillation, or recovery training
are required.

\section{Experimental Results}

\definecolor{oursbg}{RGB}{232,240,254}
\definecolor{densebg}{RGB}{243,243,243}
\newcommand{\std}[1]{\,$\pm$\,#1}
\newcommand{\zero}{\multicolumn{1}{c}{0.0}}

\begin{table*}[!t]
\centering
\scriptsize
\setlength{\tabcolsep}{2.4pt}
\renewcommand{\arraystretch}{1.08}
\begin{adjustbox}{max width=\textwidth}
\begin{tabular}{@{}l c rrrrrr rrrrrr@{}}
\toprule
& & \multicolumn{6}{c}{$L_p/L_t = 7/40$} & \multicolumn{6}{c}{$L_p/L_t = 9/40$} \\
\cmidrule(l{2pt}r{2pt}){3-8}
\cmidrule(l{2pt}r{2pt}){9-14}
Method & Type & AIME25 & AIME26 & GPQA-D & HumanEval & MBPP+ & Avg. & AIME25 & AIME26 & GPQA-D & HumanEval & MBPP+ & Avg. \\
\midrule

\rowcolor{densebg}
\multicolumn{2}{@{}l}{\textit{Qwen3-14B} \scriptsize(Dense)} & 72.2\std{5.1} & 75.6\std{10.2} & 62.3\std{0.6} & 93.5\std{0.9} & 80.0\std{1.3} & 76.7 & 72.2\std{5.1} & 75.6\std{10.2} & 62.3\std{0.6} & 93.5\std{0.9} & 80.0\std{1.3} & 76.7 \\
\cmidrule(l{2pt}r{2pt}){1-14}

ShortGPT & Depth & \zero & \zero & 5.6\std{0.5} & 19.5\std{1.2} & 6.0\std{0.6} & 6.2 & \zero & \zero & 0.3\std{0.6} & 1.4\std{0.4} & 0.2\std{0.2} & 0.4 \\
LaCo & Depth & \zero & \zero & \zero & \zero & \zero & 0.0 & \zero & \zero & \zero & \zero & \zero & 0.0 \\
GRASP & Depth & \zero & \zero & 0.5\std{0.0} & \zero & \zero & 0.1 & \zero & \zero & \zero & \zero & \zero & 0.0 \\
FLAP & Width & 44.4\std{3.8} & 51.1\std{8.4} & 44.1\std{0.8} & 76.8\std{3.2} & 51.6\std{0.8} & 53.6 & 38.9\std{6.9} & 41.1\std{6.9} & 35.4\std{3.9} & 59.8\std{2.8} & 49.3\std{0.6} & 44.9 \\
RESP & Width & 57.8\std{8.4} & \textbf{62.2}\std{3.8} & 48.0\std{2.2} & 84.3\std{2.7} & 72.8\std{1.9} & 65.0 & 52.2\std{5.1} & \textbf{67.8}\std{11.7} & \textbf{46.0}\std{2.2} & 62.0\std{2.1} & 63.7\std{1.1} & 58.3 \\
CoCurve & Width & 1.1\std{1.9} & 5.6\std{1.9} & 24.6\std{1.5} & 53.9\std{2.1} & 50.4\std{0.7} & 27.1 & \zero & \zero & 7.7\std{0.6} & 25.0\std{0.6} & 34.0\std{1.4} & 13.3 \\
\rowcolor{oursbg}
\textbf{LayerTrim (Ours)} & Width & \textbf{64.4}\std{5.1} & 60.0\std{5.8} & \textbf{49.7}\std{2.9} & \textbf{90.2}\std{1.6} & \textbf{77.2}\std{1.8} & \textbf{68.3} & \textbf{56.7}\std{3.3} & 63.3\std{3.3} & 42.8\std{3.5} & \textbf{88.0}\std{2.7} & \textbf{74.8}\std{0.3} & \textbf{65.1} \\

\midrule

\rowcolor{densebg}
\multicolumn{2}{@{}l}{\textit{Phi-4-reasoning-plus} \scriptsize(Dense)} & 72.2\std{5.1} & 67.8\std{5.1} & 67.7\std{1.7} & 78.7\std{2.4} & 68.8\std{2.2} & 71.0 & 72.2\std{5.1} & 67.8\std{5.1} & 67.7\std{1.7} & 78.7\std{2.4} & 68.8\std{2.2} & 71.0 \\
\cmidrule(l{2pt}r{2pt}){1-14}

ShortGPT & Depth & \zero & \zero & 3.0\std{0.5} & 2.8\std{1.4} & 5.4\std{1.1} & 2.2 & \zero & \zero & 9.1\std{1.3} & \zero & \zero & 1.8 \\
LaCo & Depth & \zero & \zero & \zero & \zero & \zero & 0.0 & \zero & \zero & \zero & \zero & \zero & 0.0 \\
GRASP & Depth & \zero & \zero & 0.5\std{0.5} & \zero & \zero & 0.1 & \zero & \zero & \zero & \zero & \zero & 0.0 \\
FLAP & Width & 22.2\std{1.9} & 27.8\std{5.1} & 41.4\std{1.3} & 68.9\std{5.8} & 51.5\std{0.7} & 42.4 & 7.8\std{3.8} & 7.8\std{1.9} & 24.1\std{0.8} & 3.7\std{0.0} & 2.5\std{0.9} & 9.2 \\
RESP & Width & \textbf{62.2}\std{1.9} & 50.0\std{5.8} & 33.5\std{0.6} & 50.6\std{2.1} & 53.2\std{0.9} & 49.9 & 40.0\std{5.8} & 45.6\std{5.1} & 15.7\std{0.5} & 34.1\std{1.2} & 37.6\std{1.1} & 34.6 \\
CoCurve & Width & 8.9\std{3.8} & 3.3\std{3.3} & 17.0\std{0.8} & 57.3\std{2.2} & 45.9\std{1.3} & 26.5 & 3.3\std{0.0} & 3.3\std{0.0} & 6.2\std{0.3} & 26.6\std{3.0} & 24.9\std{1.6} & 12.9 \\
\rowcolor{oursbg}
\textbf{LayerTrim (Ours)} & Width & 60.0\std{3.3} & \textbf{60.0}\std{3.3} & \textbf{53.0}\std{0.5} & \textbf{74.6}\std{3.0} & \textbf{63.3}\std{1.0} & \textbf{62.2} & \textbf{43.3}\std{3.3} & \textbf{46.7}\std{0.0} & \textbf{43.3}\std{0.6} & \textbf{56.9}\std{2.0} & \textbf{54.3}\std{1.4} & \textbf{48.9} \\

\midrule

\rowcolor{densebg}
\multicolumn{2}{@{}l}{\textit{Magistral-Small-2506} \scriptsize(Dense)} & 56.7\std{3.3} & 61.1\std{8.4} & 62.3\std{3.0} & 94.3\std{0.9} & 77.2\std{1.1} & 70.3 & 56.7\std{3.3} & 61.1\std{8.4} & 62.3\std{3.0} & 94.3\std{0.9} & 77.2\std{1.1} & 70.3 \\
\cmidrule(l{2pt}r{2pt}){1-14}

ShortGPT & Depth & \zero & \zero & \zero & \zero & \zero & 0.0 & \zero & \zero & \zero & \zero & \zero & 0.0 \\
LaCo & Depth & \zero & \zero & \zero & \zero & \zero & 0.0 & \zero & \zero & \zero & \zero & \zero & 0.0 \\
GRASP & Depth & \zero & \zero & \zero & \zero & \zero & 0.0 & \zero & \zero & \zero & \zero & \zero & 0.0 \\
FLAP & Width & 3.3\std{0.0} & 2.2\std{1.9} & 21.9\std{0.8} & 40.0\std{2.5} & 36.5\std{2.3} & 20.8 & 1.1\std{1.9} & \zero & 12.6\std{2.7} & 21.5\std{0.7} & 17.8\std{0.9} & 10.6 \\
RESP & Width & \textbf{40.0}\std{3.3} & 37.8\std{5.1} & 38.4\std{1.3} & 53.7\std{1.8} & 51.3\std{1.6} & 44.2 & \textbf{18.9}\std{6.9} & 15.6\std{6.9} & 24.1\std{2.6} & 27.6\std{0.9} & 35.5\std{0.9} & 24.3 \\
CoCurve & Width & 1.1\std{1.9} & 1.1\std{1.9} & 15.8\std{2.0} & 56.3\std{0.9} & 50.7\std{2.1} & 25.0 & \zero & \zero & 3.7\std{0.8} & 19.5\std{1.8} & 25.0\std{1.2} & 9.6 \\
\rowcolor{oursbg}
\textbf{LayerTrim (Ours)} & Width & 30.0\std{8.8} & \textbf{42.2}\std{1.9} & \textbf{52.2}\std{1.5} & \textbf{90.7}\std{3.7} & \textbf{73.8}\std{1.2} & \textbf{57.8} & \textbf{18.9}\std{8.4} & \textbf{18.9}\std{1.9} & \textbf{44.3}\std{1.6} & \textbf{76.0}\std{3.1} & \textbf{70.5}\std{2.5} & \textbf{45.7} \\

\bottomrule
\end{tabular}
\end{adjustbox}

\caption{
Main results across three RLMs under matched parameter
budgets. The two column groups correspond to $L_p/L_t=7/40$ and $9/40$,
where $L_p$ is the layer-equivalent pruning budget and $L_t$ is the
original decoder depth. Type indicates depth or width pruning.
Benchmark entries are mean\,$\pm$\,sample std.\ over three seeds,
and Avg.\ is the unweighted mean accuracy (\%) across the five benchmarks.
\textbf{Bold} marks the best pruned-model accuracy within each budget
for each benchmark and the average.
}
\label{tab:main_results}
\label{tab:main}
\end{table*}


\subsection{Experimental Settings}
\label{sec:protocol}

\paragraph{Models and benchmarks.}

We use three RLMs and five benchmarks:
\begin{itemize}[leftmargin=*]
    \item Models: \texttt{Qwen3-14B}~\citep{yang2025qwen3}, \texttt{Phi-4-reasoning-plus}~\citep{phi4reasoning}, \texttt{Magistral-Small-2506}~\citep{rastogi2025magistral};
    \item Benchmarks: AIME2025~\cite{aime25}, AIME2026~\cite{aime26}, GPQA-Diamond~\cite{rein2023gpqa}, HumanEval~\cite{humaneval}, and MBPP+~\citep{mbpp}.
\end{itemize}
We report exact numerical-answer accuracy on AIME, multiple-choice
accuracy on GPQA-Diamond, and pass@1 on HumanEval and MBPP+ using
their original and extended test suites, respectively.
For each model, we use its original dense checkpoint as the reference
to measure performance retention after pruning.

\paragraph{Baselines and pruning budgets.}

We compare LayerTrim with three depth pruning methods and three width pruning methods:
\begin{itemize}[leftmargin=*]
    \item Depth: ShortGPT \citep{men2024shortgpt}, LaCo \citep{yang2024laco}, GRASP \citep{liu2025grasp};
    \item Width: FLAP \citep{an2024flap}, RESP \citep{wang2025resp}, CoCurve \citep{gong2026cocurve}.
\end{itemize}
We evaluate budgets of $N \in \{7,9\}$ layer-equivalents,
each corresponding to the parameter count of $N$ decoder layers.
We match the pruning parameter budget across methods.
Budget allocation and method-specific implementation details
are provided in Appendix~\ref{app:conversion}.

\paragraph{Calibration.}
For each model, we use the same calibration set across all pruning methods,
consisting of 24 AIME 2024 reasoning traces generated by its dense checkpoint; following its original procedure, RESP additionally regenerates these traces during pruning.
Traces are generated with temperature $0.6$, seed $0$, and a maximum
generation length of 16K tokens.
LayerTrim collects layer redundancy and MLP feature importance statistics
from each complete prompt and up to its first 2K generated tokens,
without updating its weights.

\paragraph{Evaluation settings.}
We evaluate dense and pruned models using their native chat templates
with reasoning enabled.
All models are evaluated with temperature $0.6$ and top-$p$ $0.95$.
The effective maximum output length is 32K tokens for all three models.
We evaluate each configuration with generation seeds $0$, $1$, and $2$,
generating one response per problem for each seed.
We report the mean and sample standard deviation over the three seeds.

\subsection{Main Results}
\label{sec:results}

Table~\ref{tab:main} compares LayerTrim with existing pruning methods
under matched parameter budgets.
LayerTrim achieves the highest average benchmark score among the pruning
methods for all three models at both budgets.
Its lead over the strongest baseline grows as the budget increases
from $N=7$ to $N=9$.
At $N=9$, LayerTrim exceeds the strongest baseline average by 6.8,
14.3, and 21.4 percentage points on \texttt{Qwen3-14B},
\texttt{Phi-4-reasoning-plus}, and \texttt{Magistral-Small-2506}, respectively.

Although calibration uses only AIME 2024 reasoning traces, LayerTrim's
advantages extend to science and coding.
Compared with the strongest baseline on each benchmark,
its gains reach up to 20.2 percentage points on GPQA-Diamond,
48.4 on HumanEval, and 35.0 on MBPP+.

\subsection{Ablation Studies}
\label{sec:ablations}

We examine four design choices in LayerTrim: the pruning target,
feature-importance score, layer selection, and number of selected MLPs.
Table~\ref{tab:ablations} reports results on \texttt{Qwen3-14B} using AIME25
and GPQA-Diamond. Detailed experimental settings are provided in
Appendix~\ref{app:ablation-details}.
Additional calibration-length, calibration-source, and cross-feature
selection ablations appear in Appendix~\ref{app:additional-accuracy-ablations}.

\begin{table}[!t]
\centering
\scriptsize
\setlength{\tabcolsep}{1.5pt}
\renewcommand{\arraystretch}{1.15}

\begin{tabular*}{\linewidth}{
@{\extracolsep{\fill}}
>{\raggedright\arraybackslash}p{0.40\linewidth}
>{\centering\arraybackslash}p{0.21\linewidth}
>{\centering\arraybackslash}p{0.21\linewidth}
>{\centering\arraybackslash}p{0.12\linewidth}
@{}}
\toprule
\textbf{Variant} &
\textbf{AIME25} &
\textbf{GPQA-D} &
\textbf{Avg.} \\
\midrule

\rowcolor{black!7}
\multicolumn{4}{@{}l@{}}{
\textbf{Pruning targets}
} \\
\addlinespace[2pt]
Attention removal
& 0.0\std{0.0} & 0.7\std{0.3} & 0.4 \\
Attention trimming
& 0.0\std{0.0} & 1.3\std{0.3} & 0.7 \\
MLP removal
& 0.0\std{0.0} & 17.5\std{1.1} & 8.8 \\
Joint trimming (25\% attn.)
& 43.3\std{3.3} & 53.2\std{4.2} & 48.3 \\
\rowcolor{oursbg}
MLP trimming
& 62.2\std{8.4} & 55.4\std{2.4} & \textbf{58.8} \\

\addlinespace[3pt]

\rowcolor{black!7}
\multicolumn{4}{@{}l@{}}{
\textbf{Feature-importance score}
} \\
\addlinespace[2pt]
$\mathbb{E}[\phi_j^2]$
& 61.1\std{5.1} & 48.1\std{3.3} & 54.6 \\
$\|w_j\|_2^2$
& 22.2\std{1.9} & 41.1\std{3.0} & 31.7 \\
$\operatorname{Var}(\phi_j)\|w_j\|_2^2$ (FLAP)
& 60.0\std{8.8} & 49.5\std{4.5} & 54.8 \\
\rowcolor{oursbg}
$\mathbb{E}[\phi_j^2]\|w_j\|_2^2$ (Ours)
& 64.4\std{5.1} & 49.7\std{2.9} & \textbf{57.1} \\

\addlinespace[3pt]

\rowcolor{black!7}
\multicolumn{4}{@{}l@{}}{
\textbf{Layer selection and distribution}
} \\
\addlinespace[2pt]
MLP-only BI
& 56.7\std{5.8} & 51.2\std{1.5} & 54.0 \\
Random
& 50.0\std{8.8} & 46.3\std{0.8} & 48.2 \\
All-layer ($K=40$)
& 50.0\std{8.8} & 53.5\std{0.9} & 51.8 \\
\rowcolor{oursbg}
Block BI (Ours)
& 64.4\std{5.1} & 49.7\std{2.9} & \textbf{57.1} \\

\addlinespace[3pt]

\rowcolor{black!7}
\multicolumn{4}{@{}l@{}}{
\textbf{Number of selected MLPs}
} \\
\addlinespace[2pt]
$K=1.5N$
& 66.7\std{3.3} & 46.6\std{5.4} & 56.7 \\
$K=3N$
& 57.8\std{3.8} & 54.9\std{1.5} & 56.4 \\
\rowcolor{oursbg}
$K=2N$ (Ours)
& 64.4\std{5.1} & 49.7\std{2.9} & \textbf{57.1} \\

\bottomrule
\end{tabular*}

\caption{
Ablations on \texttt{Qwen3-14B}. Scores (\%) are mean $\pm$ sample std
over three seeds. Avg. averages AIME25 and GPQA-Diamond;
bold marks the best average in each group.
We remove approximately 1.5B parameters in the first group
and 2.312B ($N=7$) in the others.
See Appendix~\ref{app:ablation-details} for experimental settings.
}
\label{tab:ablations}
\end{table}

\paragraph{Pruning targets.}
Figure~\ref{fig:layertrim_observations}(b) shows that MLP trimming
produces the lowest final-hidden NMSE.
The accuracy results in Table~\ref{tab:ablations} show the same pattern: MLP-only trimming preserves
substantially higher average accuracy than whole-module removal or
attention pruning. It also outperforms joint attention–MLP trimming, which allocates 25\% of the pruning budget to attention.
These results show that MLP-only trimming minimizes pruning
damage among the tested choices in both representation error and
task accuracy.

\paragraph{Feature-Importance score.} We compare alternative criteria for selecting MLP features during width reduction. Scores based on activations achieve higher average accuracy than weight-only ranking. Our feature-importance score achieves the highest average accuracy among the evaluated criteria, outperforming activation-only ranking and FLAP's variance-based score.

\paragraph{Layer selection and distribution.}
The accuracy results in Table~\ref{tab:ablations} support the BI-based allocation in
Figure~\ref{fig:layertrim_observations}(d): selecting low-BI layers
outperforms random selection and all-layer trimming in average accuracy.
Block BI also achieves higher average accuracy than MLP-only BI,
which measures changes in the residual stream before and after the MLP update.
This supports measuring redundancy across the full decoder block
rather than the MLP update alone.

\paragraph{Number of selected MLPs.}
Under a fixed budget, $K=1.5N$ achieves the highest AIME25 accuracy,
while $K=3N$ achieves the highest GPQA-Diamond accuracy.
The default $K=2N$ gives the highest average accuracy by a small margin,
providing a practical balance between the two tasks.

\renewcommand{\tracecell}[1]{%
  \begin{minipage}[t]{0.235\textwidth}\raggedright #1\end{minipage}}

\begin{table*}[!t]
\centering
\scriptsize
\setlength{\tabcolsep}{2.6pt}
\renewcommand{\arraystretch}{1.08}
\begin{adjustbox}{max width=\textwidth}
\begin{tabular}{@{}l c cccr cccr@{}}
\toprule
& & \multicolumn{4}{c}{$L_p/L_t = 7/40$} & \multicolumn{4}{c}{$L_p/L_t = 9/40$} \\
\cmidrule(l{2pt}r{2pt}){3-6}
\cmidrule(l{2pt}r{2pt}){7-10}
Method & Type & Prefill (ms) & Decode (ms/tok) & Peak Mem.\ (GiB) & Acc.\ (\%) & Prefill (ms) & Decode (ms/tok) & Peak Mem.\ (GiB) & Acc.\ (\%) \\
\midrule

\rowcolor{densebg}
\multicolumn{2}{@{}l}{\textit{Qwen3-14B} \scriptsize(Dense)}
& 588 (1.00$\times$) & 38.0 (1.00$\times$) & 28.1 (100\%) & 76.7 (100.0\%)
& 588 (1.00$\times$) & 38.0 (1.00$\times$) & 28.1 (100\%) & 76.7 (100.0\%) \\
\cmidrule(l{2pt}r{2pt}){1-10}

ShortGPT & Depth
& 502 (1.17$\times$) & 31.8 (1.19$\times$) & 23.8 (84\%) & 6.2 (8.1\%)
& 470 (1.25$\times$) & 30.1 (1.26$\times$) & 22.5 (80\%) & 0.4 (0.5\%) \\

LaCo & Depth
& 497 (1.18$\times$) & 31.8 (1.19$\times$) & 23.8 (84\%) & 0.0 (0.0\%)
& 466 (1.26$\times$) & 30.1 (1.26$\times$) & 22.5 (80\%) & 0.0 (0.0\%) \\

GRASP & Depth
& 498 (1.18$\times$) & 34.0 (1.12$\times$) & 23.8 (85\%) & 0.1 (0.1\%)
& 466 (1.26$\times$) & 32.8 (1.16$\times$) & 22.6 (80\%) & 0.0 (0.0\%) \\

FLAP & Width
& 513 (1.15$\times$) & 32.6 (1.16$\times$) & 23.6 (84\%) & 53.6 (69.9\%)
& 482 (1.22$\times$) & 31.2 (1.22$\times$) & 22.4 (80\%) & 44.9 (58.5\%) \\

RESP & Width
& 523 (1.12$\times$) & 32.9 (1.15$\times$) & 23.8 (85\%) & 65.0 (84.7\%)
& 493 (1.19$\times$) & 31.4 (1.21$\times$) & 22.6 (80\%) & 58.3 (76.0\%) \\

CoCurve & Width
& 509 (1.16$\times$) & 32.8 (1.16$\times$) & 23.8 (85\%) & 27.1 (35.3\%)
& 508 (1.16$\times$) & 31.4 (1.21$\times$) & 22.5 (80\%) & 13.3 (17.3\%) \\

\rowcolor{oursbg}
LayerTrim (Ours) & Width
& 510 (1.15$\times$) & 32.8 (1.16$\times$) & 23.8 (85\%) & 68.3 (89.0\%)
& 474 (1.24$\times$) & 31.3 (1.21$\times$) & 22.6 (80\%) & 65.1 (84.9\%) \\

\midrule

\rowcolor{densebg}
\multicolumn{2}{@{}l}{\textit{Phi-4-reasoning-plus} \scriptsize(Dense)}
& 635 (1.00$\times$) & 37.7 (1.00$\times$) & 28.2 (100\%) & 71.0 (100.0\%)
& 635 (1.00$\times$) & 37.7 (1.00$\times$) & 28.2 (100\%) & 71.0 (100.0\%) \\
\cmidrule(l{2pt}r{2pt}){1-10}

ShortGPT & Depth
& 539 (1.18$\times$) & 31.5 (1.20$\times$) & 23.6 (84\%) & 2.2 (3.1\%)
& 497 (1.28$\times$) & 29.8 (1.27$\times$) & 22.3 (79\%) & 1.8 (2.5\%) \\

LaCo & Depth
& 540 (1.17$\times$) & 31.5 (1.20$\times$) & 23.6 (84\%) & 0.0 (0.0\%)
& 505 (1.26$\times$) & 29.8 (1.27$\times$) & 22.3 (79\%) & 0.0 (0.0\%) \\

GRASP & Depth
& 548 (1.16$\times$) & 32.6 (1.16$\times$) & 23.7 (84\%) & 0.1 (0.1\%)
& 514 (1.23$\times$) & 31.1 (1.21$\times$) & 22.4 (80\%) & 0.0 (0.0\%) \\

FLAP & Width
& 521 (1.22$\times$) & 32.4 (1.16$\times$) & 23.5 (83\%) & 42.4 (59.7\%)
& 498 (1.27$\times$) & 30.8 (1.22$\times$) & 22.2 (79\%) & 9.2 (13.0\%) \\

RESP & Width
& 544 (1.17$\times$) & 32.4 (1.16$\times$) & 23.7 (84\%) & 49.9 (70.3\%)
& 514 (1.23$\times$) & 30.9 (1.22$\times$) & 22.4 (80\%) & 34.6 (48.7\%) \\

CoCurve & Width
& 533 (1.19$\times$) & 32.6 (1.15$\times$) & 23.6 (84\%) & 26.5 (37.3\%)
& 515 (1.23$\times$) & 30.9 (1.22$\times$) & 22.3 (79\%) & 12.9 (18.2\%) \\

\rowcolor{oursbg}
LayerTrim (Ours) & Width
& 557 (1.14$\times$) & 32.3 (1.17$\times$) & 23.7 (84\%) & 62.2 (87.6\%)
& 519 (1.22$\times$) & 30.8 (1.22$\times$) & 22.4 (80\%) & 48.9 (68.9\%) \\

\midrule

\rowcolor{densebg}
\multicolumn{2}{@{}l}{\textit{Magistral-Small-2506} \scriptsize(Dense)}
& 996 (1.00$\times$) & 57.4 (1.00$\times$) & 44.7 (100\%) & 70.3 (100.0\%)
& 996 (1.00$\times$) & 57.4 (1.00$\times$) & 44.7 (100\%) & 70.3 (100.0\%) \\
\cmidrule(l{2pt}r{2pt}){1-10}

ShortGPT & Depth
& 836 (1.19$\times$) & 47.9 (1.20$\times$) & 37.4 (84\%) & 0.0 (0.0\%)
& 779 (1.28$\times$) & 45.2 (1.27$\times$) & 35.3 (79\%) & 0.0 (0.0\%) \\

LaCo & Depth
& 836 (1.19$\times$) & 47.9 (1.20$\times$) & 37.4 (84\%) & 0.0 (0.0\%)
& 781 (1.28$\times$) & 45.2 (1.27$\times$) & 35.3 (79\%) & 0.0 (0.0\%) \\

GRASP & Depth
& 797 (1.25$\times$) & 49.2 (1.17$\times$) & 37.5 (84\%) & 0.0 (0.0\%)
& 763 (1.31$\times$) & 46.9 (1.23$\times$) & 35.4 (79\%) & 0.0 (0.0\%) \\

FLAP & Width
& 783 (1.27$\times$) & 48.8 (1.18$\times$) & 37.4 (84\%) & 20.8 (29.6\%)
& 789 (1.26$\times$) & 46.4 (1.24$\times$) & 35.3 (79\%) & 10.6 (15.1\%) \\

RESP & Width
& 810 (1.23$\times$) & 48.8 (1.18$\times$) & 37.5 (84\%) & 44.2 (62.9\%)
& 806 (1.24$\times$) & 46.3 (1.24$\times$) & 35.4 (79\%) & 24.3 (34.6\%) \\

CoCurve & Width
& 778 (1.28$\times$) & 48.8 (1.18$\times$) & 37.4 (84\%) & 25.0 (35.6\%)
& 789 (1.26$\times$) & 46.4 (1.24$\times$) & 35.3 (79\%) & 9.6 (13.7\%) \\

\rowcolor{oursbg}
LayerTrim (Ours) & Width
& 834 (1.19$\times$) & 48.8 (1.18$\times$) & 37.5 (84\%) & 57.8 (82.2\%)
& 781 (1.27$\times$) & 46.3 (1.24$\times$) & 35.4 (79\%) & 45.7 (65.0\%) \\

\bottomrule
\end{tabular}
\end{adjustbox}
\caption{Inference efficiency under matched parameter budgets across three RLMs. We report prefill latency, decode latency per token, peak GPU memory, and average task accuracy. Parentheses indicate speedup over the dense model for latency, relative memory usage for peak GPU memory, and accuracy retention relative to the dense model. All methods are evaluated using their deployed pruned structures.}
\label{tab:latency}
\end{table*}

\subsection{Efficiency Measurement}
\paragraph{Measurement settings.}
We measure prefill latency, decode latency, and peak GPU memory
in bfloat16 at batch size 1 on a single NVIDIA RTX 6000 Ada (48\,GB).
Prefill latency is measured on a 2{,}048-token prompt, and decode
latency is averaged over 128 greedy decoding steps with a static KV cache.
Each configuration runs two warm-up iterations followed by ten
timed runs, synchronized using \texttt{torch.cuda.synchronize()}.
We report median latency and peak GPU memory during inference,
excluding model loading and pruning overhead. All methods are evaluated under matched parameter budgets.

\paragraph{Efficiency results.}
Table~\ref{tab:latency} compares inference efficiency across
three RLMs at two pruning budgets.
At the higher pruning budget, LayerTrim achieves prefill speedups
of 1.24$\times$, 1.22$\times$, and 1.27$\times$, and decode speedups
of 1.21$\times$, 1.22$\times$, and 1.24$\times$ on \texttt{Qwen3-14B},
\texttt{Phi-4-reasoning-plus}, and \texttt{Magistral-Small-2506}, respectively.
Peak GPU memory is reduced by 19.7\%, 20.3\%, and 20.8\%
on the corresponding models compared with their dense counterparts.
While achieving inference costs comparable to existing structured
pruning methods, LayerTrim retains the highest average accuracy
across all three models at both pruning budgets.
LayerTrim also requires only 27.9 s of pruning preparation on Qwen3-14B, 94× and 140× faster than RESP at N = 7 and N = 9, respectively (Appendix ~\ref{app:calibration-overhead}).

\section{Discussions}
\subsection{Analysis of LayerTrim}
\label{sec:analysis}

\begin{figure*}[htb!]
    \centering
    \begin{subfigure}[t]{.6\linewidth}
        \centering
        \includegraphics[width=\linewidth]{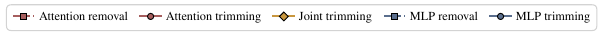}
    \end{subfigure}

    \begin{subfigure}[t]{.32\linewidth}
        \centering
        \includegraphics[width=\linewidth]{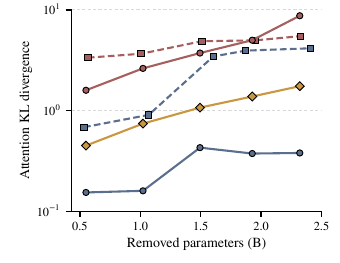}
        \caption{Attention distribution}
        \label{fig:attention-kl-panel}
    \end{subfigure}
    \hfill
    \begin{subfigure}[t]{.32\linewidth}
        \centering
        \includegraphics[width=\linewidth]{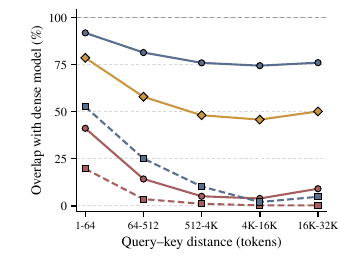}
        \caption{Attention overlap}
        \label{fig:attention-overlap-panel}
    \end{subfigure}
    \hfill
    \begin{subfigure}[t]{.32\linewidth}
        \centering
        \includegraphics[width=\linewidth]{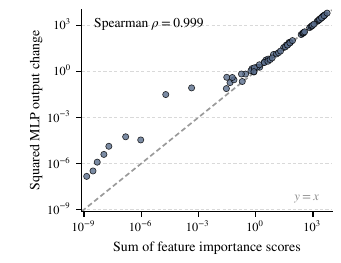}
        \caption{Feature importance}
        \label{fig:feature-importance-surrogate-panel}
    \end{subfigure}
    \caption{Attention preservation and feature importance on \texttt{Qwen3-14B}.
    (a) Attention KL divergence across pruning budgets.
    (b) Overlap with the dense model's top-8 attended tokens by query--key distance,
    with approximately 1.5B parameters removed.
    (c) Summed feature importance scores versus squared MLP output change.
    Experimental details are provided in Appendix~\ref{app:attention-surrogate-analysis}.}
    \label{fig:attention-surrogate-analysis}
\end{figure*}

\paragraph{Attention preservation.} We analyze how well MLP trimming preserves attention patterns on reasoning traces. Panel~\ref{fig:attention-kl-panel} shows that MLP trimming produces the lowest attention KL divergence across the tested budgets. At approximately 1.5B removed parameters, panel~\ref{fig:attention-overlap-panel} also shows the highest overlap with the dense model's top-8 attended tokens across all query--key distance bins, including 16K--32K. These results suggest that MLP-only trimming better preserves attention to earlier tokens during long reasoning, providing an attention-level explanation for why MLP trimming better preserves reasoning performance.

\paragraph{Feature-Importance Surrogate.} To reduce calibration overhead, we approximate the squared MLP output change by summing feature importance scores without computing cross-feature terms. For a set $R$ of removed features, we compare this approximation with the squared output change measured on the same dense-model inputs:
\begin{equation}
    \begin{aligned}
    \Delta_\ell(R)
    &=\mathbb{E}_{x}\!\left[
      \left\|\sum_{j\in R}c_j^{(\ell)}(x)\right\|_2^2\right]
    &\approx\sum_{j\in R}s_j^{(\ell)}
    \end{aligned}
    \label{eq:analysis_mlp_output_change}
\end{equation}

Panel~\ref{fig:feature-importance-surrogate-panel} shows that the scores rank the measured changes across held-out layer--width conditions, with Spearman $\rho=0.999$. Table~\ref{tab:cross-term-accuracy} in Appendix~\ref{app:additional-accuracy-ablations} reports comparable average accuracy to full-error greedy selection with 13.2$\times$ faster calibration and selection. These results support using the low-overhead approximation to guide feature selection.

\subsection{Quantitative Analysis of Reasoning Traces}

\begin{figure*}[htb!]
    \centering
    \begin{subfigure}[t]{.75\linewidth}
        \centering
        \includegraphics[width=\linewidth]{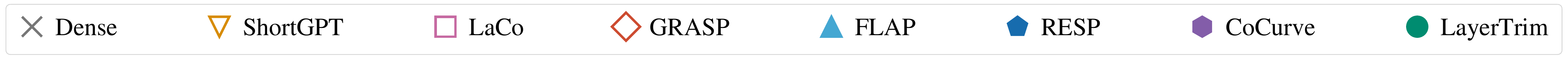}
    \end{subfigure}

    \begin{subfigure}[t]{.32\linewidth}
        \centering
        \includegraphics[width=\linewidth]{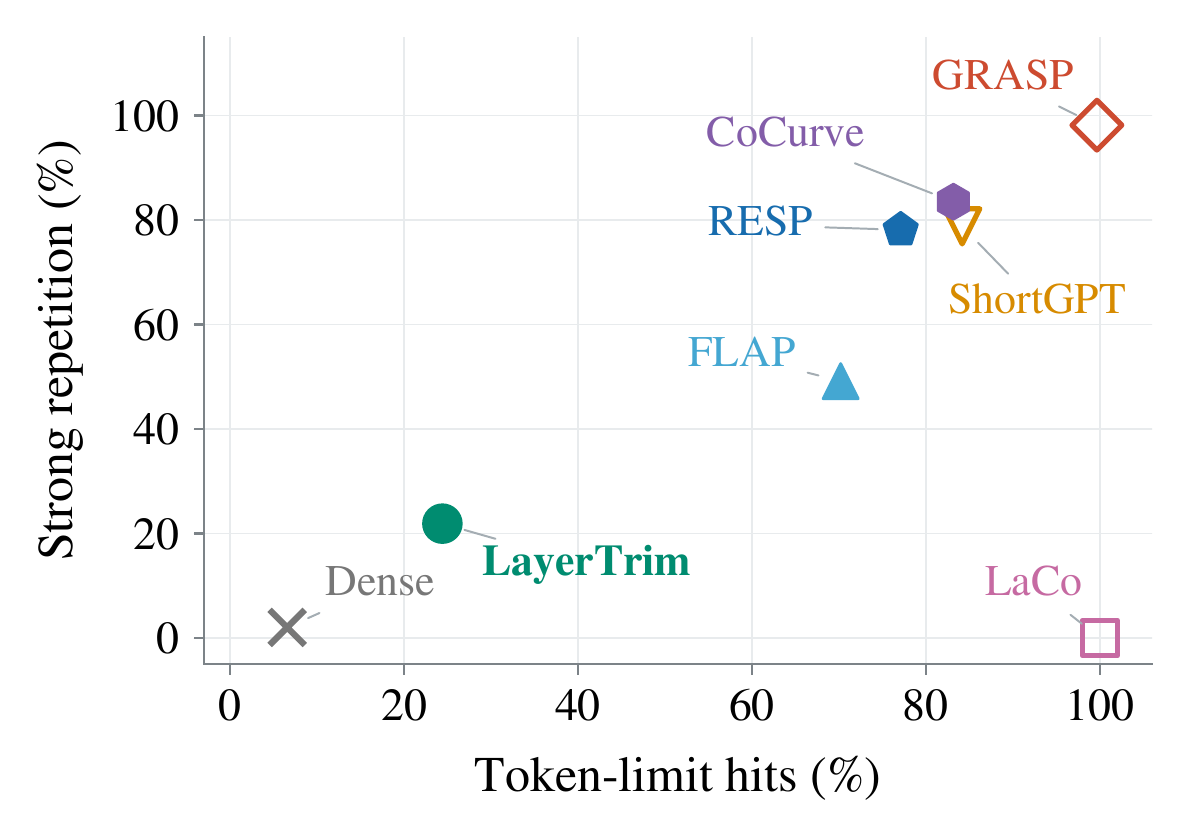}
        \caption{Repetition, GPQA-D}
        \label{fig:gpqa_repitition}
    \end{subfigure}
    \hfill
    \begin{subfigure}[t]{.32\linewidth}
        \centering
        \includegraphics[width=\linewidth]{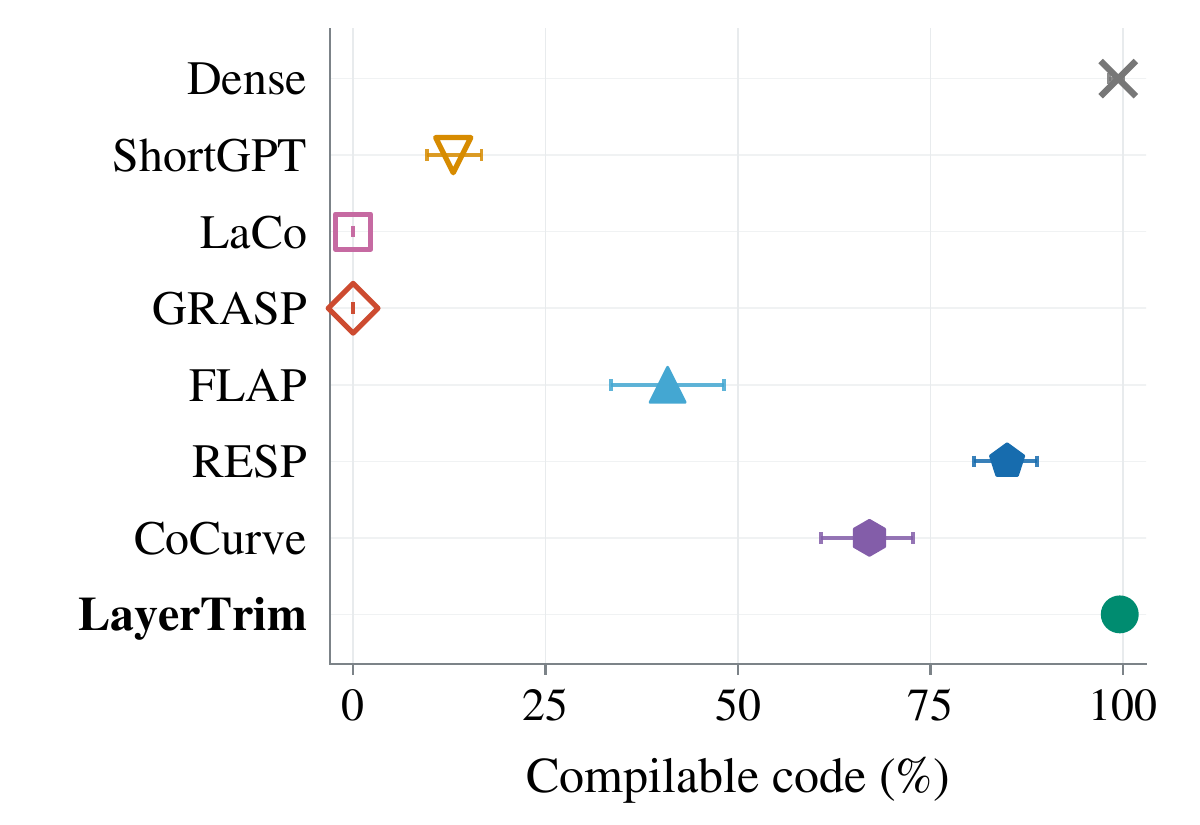}
        \caption{Coding, HumanEval}
        \label{fig:coding_capability}
    \end{subfigure}
    \hfill
    \begin{subfigure}[t]{.32\linewidth}
        \centering
        \includegraphics[width=\linewidth]{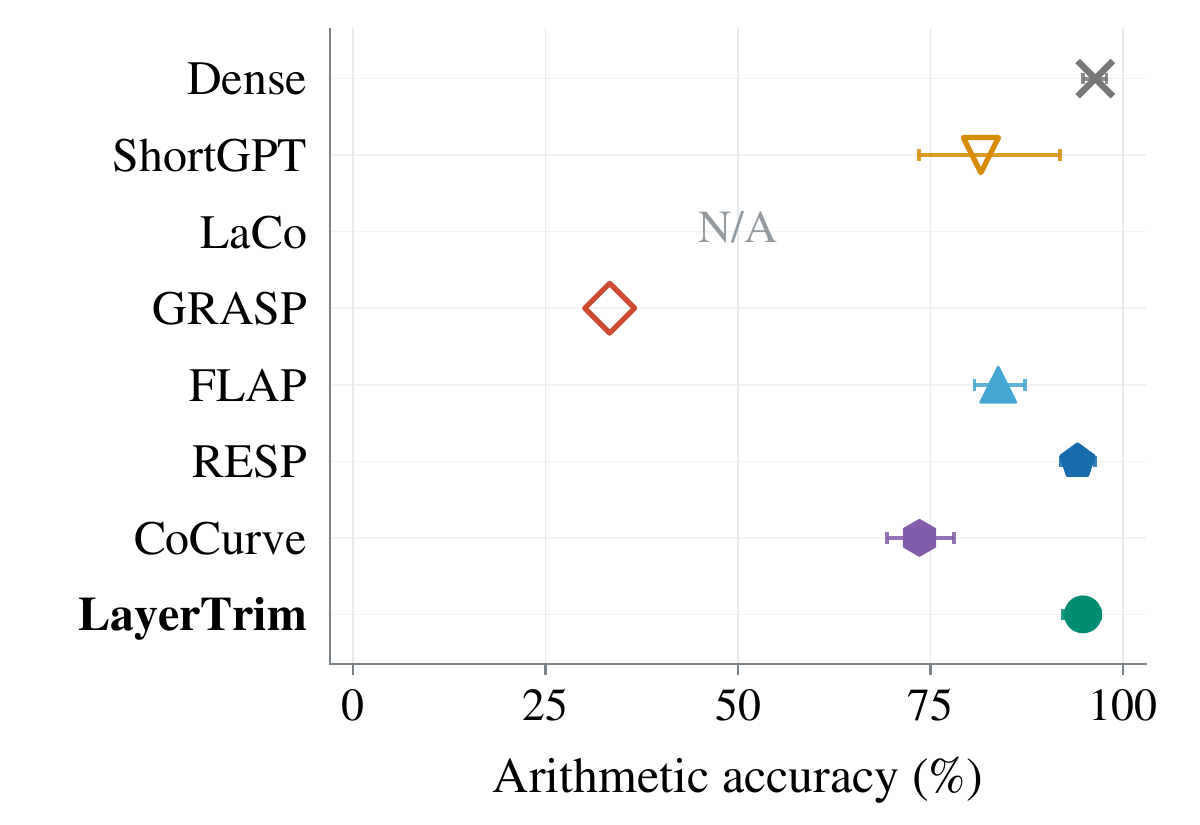}
        \caption{Arithmetic, AIME26}
        \label{fig:arithmetic_consistency}
    \end{subfigure}
    \caption{Quantitative analysis results comparing LayerTrim to existing pruning methods}
    \label{fig:quant_analysis}
\end{figure*}

Figure~\ref{fig:quant_analysis} complements the experimental results in Table~\ref{tab:main_results} with diagnostics of the generated outputs and reasoning paths.
Panel~\ref{fig:gpqa_repitition} examines GPQA-Diamond responses from \texttt{Phi-4-reasoning-plus} at $L_p/L_t=9/40$, measuring the ratio of responses reaching the max token limit and exhibiting strong repetition.
Here, we flag strong repetition when two consecutive, non-overlapping 512-token windows each satisfy $1-U_8/N_8 \geq 0.8$, where $U_8$ and $N_8$ denote the numbers of unique and total token 8-grams.
LaCo exhibits 100\% token-limit hits despite 0\% detected strong repetition: its outputs continue as incoherent token fragments without forming the sustained exact-token repetition captured by this criterion (see Appendix~\ref{sec:qual_analysis}).
LayerTrim achieves token-limit hits and strong repetition rates of 24.4\% and 21.9\%, respectively, compared with 70.2\% and 49.2\% for FLAP.
 Together with its higher accuracy, these results show that LayerTrim combines stronger task performance with substantially fewer repetitive and budget-exhausting outputs.

Panel~\ref{fig:coding_capability} evaluates code formulation on HumanEval using \texttt{Qwen3-14B} at $L_p/L_t=9/40$.
The compilable-code rate measures the ratio of responses containing at least one extracted Python code that compiles and defines the requested function.
LayerTrim yields 99.6\%, close to Dense's 99.4\%, while the second-best RESP yields 85.0\%.
Its advantage also extends to functional problem solving: among responses containing a qualifying candidate, benchmark correctness is 88.4\% for LayerTrim vs. 72.7\% for RESP.
Thus, the overall HumanEval accuracy gap in Table~\ref{tab:main_results} (88.0\% versus 62.0\%) is accompanied by advantages in both forming compilable code and producing correct solutions when such code is present.

Panel~\ref{fig:arithmetic_consistency} examines arithmetic consistency on AIME26 using \texttt{Phi-4-reasoning-plus} at $L_p/L_t=7/40$.
We broadly extract numeric equalities, including those embedded in prose or chained expressions, deduplicate them within each response, and report the fraction that hold under exact arithmetic evaluation.
N/A in LaCo denotes that no eligible equality was extracted.
LayerTrim attains 94.8\% consistency, within 1.6pp of Dense's 96.4\%, compared with 83.8\% for FLAP and 73.6\% for CoCurve.
While RESP exhibits similar arithmetic consistency of 94.1\%, LayerTrim achieves substantially higher accuracy (60.0\% vs. 50.0\%).
This suggests that LayerTrim's advantage extends beyond preserving individual arithmetic operations to better preserving the problem-level reasoning needed to integrate locally correct calculations into a correct final solution.
\section{Conclusion}
LayerTrim reduces the inference cost of reasoning language models by
selectively trimming MLP features while preserving attention and decoder
depth. It uses dense-model reasoning traces to identify redundant layers
and retain high-importance features, without gradients or recovery training.
Across three models and five benchmarks, LayerTrim achieves higher average
accuracy than the evaluated structured pruning baselines at both pruning
budgets, while achieving comparable inference speedups.
LayerTrim offers a step toward more efficient deployment of reasoning
language models through training-free structured pruning.

\section*{Limitations}
Our evaluation focuses on three reasoning language models
and five benchmarks covering mathematical reasoning,
scientific reasoning, and code generation.
Although LayerTrim outperforms the evaluated pruning baselines
in most settings, experiments on additional model architectures,
scales, and reasoning tasks would help assess its generalizability.
We also evaluate fixed pruning budgets and layer allocation settings.
Adapting these choices to individual models and tasks may further
improve the trade-off between model compression and reasoning accuracy.

\section*{Ethical considerations}
Structured pruning modifies the internal architecture of language
models and may affect their reasoning behavior.
Although LayerTrim aims to preserve reasoning accuracy,
pruned models may still produce incorrect answers, incomplete
reasoning traces, or unexpected outputs.
These risks should be considered when deploying compressed
models in applications where reasoning accuracy is critical.
Our experiments use publicly available models and benchmarks
and do not involve human participants or the collection of
personal information.

\bibliography{references}

\clearpage
\appendix
\section*{Appendix}
\section{LayerTrim Procedure}
\label{app:layertrim-algorithm}

Algorithm~\ref{alg:layertrim} summarizes the procedure in
Section~\ref{sec:layertrim_procedure} for uniform gated MLPs.
Here, $L$ is the decoder depth, $C$ and $I$ are the hidden and
intermediate widths, and $P_{\mathrm{layer}}$ counts the attention
and MLP projection parameters in one original block.
Calibration averages are taken over token positions.

\begin{algorithm}[H]
\caption{LayerTrim: reasoning-calibrated structured MLP pruning.}
\label{alg:layertrim}
\small
\begin{algorithmic}[1]

\Require Dense model $\mathcal{M}$; calibration prompts $\mathcal{Q}$;
layer-equivalent budget $N$
\Ensure Compressed model $\widetilde{\mathcal{M}}$ with unchanged attention and depth

\State Generate reasoning traces
$\mathcal{T}\leftarrow\textsc{Generate}(\mathcal{M},\mathcal{Q})$.
\State Form $\mathcal{D}$ from each complete prompt and up to
2,048 generated tokens of its trace in $\mathcal{T}$.

\State With gradients disabled, teacher-force $\mathcal{M}$ on
$\mathcal{D}$ and accumulate block cosine similarities and
feature second moments $\widehat{a}_{j}^{(\ell)}
=\mathbb{E}_{\mathcal{D}}[(\phi_j^{(\ell)})^2]$.

\For{$\ell=1,\ldots,L$}
    \State $\mathrm{BI}_{\ell}\leftarrow
    1-\mathbb{E}_{x\sim\mathcal{D}}
    \!\left[
    \cos\!\left(
    h_{\mathrm{in}}^{(\ell)}(x),
    h_{\mathrm{out}}^{(\ell)}(x)
    \right)
    \right]$.
\EndFor

\State $K\leftarrow 2N$; require $K\leq L$.
\State
$\mathcal{L}_{K}\leftarrow
\textsc{BottomK}
(\{\mathrm{BI}_{\ell}\}_{\ell=1}^{L},K)$.

\State $m\leftarrow
\operatorname{round}
(I-P_{\mathrm{layer}}/(6C))$.
\State Require $0<m\leq I$.

\State $\widetilde{\mathcal{M}}\leftarrow\textsc{Copy}(\mathcal{M})$.

\ForAll{$\ell\in\mathcal{L}_{K}$}
    \For{$j=1,\ldots,I$}
        \State $\widehat{s}_{j}^{(\ell)}
        \leftarrow
        \widehat{a}_{j}^{(\ell)}
        \|W_{\mathrm{down}}^{(\ell)}[:,j]\|_2^2$.
    \EndFor

    \State $S_{\ell}^{\star}\leftarrow
    \textsc{TopM}
    (\{\widehat{s}_{j}^{(\ell)}\}_{j=1}^{I},m)$.

    \State $\widetilde{W}_{\mathrm{gate}}^{(\ell)}
    \leftarrow
    W_{\mathrm{gate}}^{(\ell)}[S_{\ell}^{\star},:]$;
    $\widetilde{W}_{\mathrm{up}}^{(\ell)}
    \leftarrow
    W_{\mathrm{up}}^{(\ell)}[S_{\ell}^{\star},:]$.

    \State $\widetilde{W}_{\mathrm{down}}^{(\ell)}
    \leftarrow
    W_{\mathrm{down}}^{(\ell)}[:,S_{\ell}^{\star}]$.
    \State Install these projections in
    $\widetilde{\mathcal{M}}$ and set this MLP's width to $m$.
\EndFor

\State \Return $\widetilde{\mathcal{M}}$

\end{algorithmic}
\end{algorithm}

\textsc{TopM} returns the indices of the $m$ highest-scoring features.
The same indices select gate and up rows and down-projection columns.
\section{Main Evaluation Implementation Details}
\label{app:conversion}

\paragraph{Pruning budgets.}
For each model, we set the pruning budget to $B_N=N P_{\mathrm{layer}}$,
where $P_{\mathrm{layer}}$ is the number of attention and MLP projection
parameters in one original decoder layer and $N\in\{7,9\}$.
All methods use the same target budget, with small differences in
actual removal counts because heads and channels are removed in whole units.

\paragraph{LayerTrim allocation.}
LayerTrim selects $K=2N$ MLPs using Block Influence and retains the same
width $m$ in each selected MLP.
Removing one gated-MLP feature affects a gate row, an up row, and a
down-projection column, totaling $3C$ parameters.
The realized budget is $3CK(I-m)$, with $m$ set by
Equation~\ref{eq:keep_width}.

\paragraph{Calibration settings.}
All methods use the same 24 AIME 2024 reasoning traces
generated by the corresponding dense model at temperature $0.6$,
top-$p$ $0.95$, and seed 0.
LayerTrim uses the complete prompt and up to the first 2,048 generated
tokens from each trace for statistics collection.

\paragraph{Baseline implementations.}
For all baselines except RESP, we use the authors' official implementations,
adapted to our models and matched-budget.
We reimplement RESP using the public GISP~\citep{gisp}  repository, following the
GISP-based structural pruning procedure described in the RESP paper
\citep{wang2025resp}, with decode-only scoring and calibration regeneration.
All baselines use the shared calibration data described above.

\paragraph{Generation settings.}
Dense and pruned models use the corresponding native reasoning-enabled
chat template. Magistral uses its native Tekken tokenizer and official
system prompt.
Sampling uses temperature $0.6$, top-$p$ $0.95$, and seeds 0, 1, and 2.
Each seed produces one response per problem.
We set the maximum output length to 32K tokens.
Inference uses BF16 in vLLM. 

\paragraph{Answer and code scoring.}
We evaluate AIME and GPQA-Diamond by exact matching of the extracted
final answers.
For coding benchmarks, we report pass@1 using HumanEval's original
tests and MBPP+'s extended tests.

\subsection{Calibration Overhead}
\label{app:calibration-overhead}

We compare LayerTrim and RESP on \texttt{Qwen3-14B} in BF16
using a single NVIDIA RTX 6000 Ada (48\,GB).
Both use the same 24 AIME 2024 calibration traces, with up to
2,048 generated tokens per trace.
We measure calibration time from model loading until the pruning masks are computed.
RESP's trace regeneration and required intermediate checkpoint
preparation are included.
Initial trace generation and final weight export are excluded.

\begin{table}[t]
\centering
\begin{adjustbox}{max width=\linewidth}
\begin{tabular}{@{}lrr@{}}
\toprule
Method & $N=7$ & $N=9$ \\
\midrule
RESP & 2,626.1\std{18.2} & 3,898.1\std{18.0} \\
\rowcolor{oursbg}
LayerTrim (Ours) & \textbf{27.9}\std{0.6} & \textbf{27.9}\std{0.6} \\
\bottomrule
\end{tabular}
\end{adjustbox}
\caption{Pruning preparation time (seconds) on Qwen3-14B.
Values are mean $\pm$ sample std over three runs.}
\label{tab:calibration-overhead}
\end{table}

LayerTrim requires substantially less preparation time than RESP
at both pruning budgets (Table~\ref{tab:calibration-overhead}).

\section{Ablation Experimental Details}
\label{app:ablation-details}

\paragraph{Evaluation and reporting.}
Table~\ref{tab:ablations} evaluates \texttt{Qwen3-14B} on AIME25 and
GPQA-Diamond using the main evaluation protocol.
The repeated default entries reuse the same evaluated configuration.

\paragraph{Pruning targets.}
The first group uses the pruning configurations of
Figure~\ref{fig:layertrim_observations}(b).
All three trimming conditions remove approximately 1.494B parameters;
attention and MLP removal delete approximately 1.510B and 1.604B,
respectively, because whole-module removal has discrete budgets.
Joint trimming assigns 25\% of its budget to attention and 75\% to MLPs.
Attention trimming removes selected dimensions within each attention head.

\paragraph{Design-ablation budget and calibration.}
The remaining three groups fix the budget at $N=7$ (approximately
2.312B removed parameters) and use the default reasoning-trace
calibration in Section~\ref{sec:protocol}.

\paragraph{Feature-importance scores.}
The feature-score group fixes the 14 Block-BI-selected MLPs
and their retained widths. Only the within-MLP ranking score changes.
Here, $\phi_j$ denotes an intermediate activation and
$w_j=W_{\mathrm{down}}[:,j]$ denotes its down-projection column.
The default score is $\mathbb{E}[\phi_j^2]\|w_j\|_2^2$.
We compare it with activation-only ranking using
$\mathbb{E}[\phi_j^2]$, weight-only ranking using $\|w_j\|_2^2$,
and variance-based ranking using
$\operatorname{Var}(\phi_j)\|w_j\|_2^2$.

\paragraph{Layer selection and distribution.}
The layer-selection group fixes the feature score and total
removal budget. Block BI, MLP-only BI, and random
selection each choose $K=14$ MLPs.
MLP-only BI measures cosine distance between the residual stream
before and after the MLP update.
Random selection uses one random fixed mask.
All-layer allocation distributes the same removal budget across
all 40 MLPs.

\paragraph{Number of selected MLPs.}
The allocation group fixes the Block BI ranking and default
feature score while varying $K\in\{11,14,21\}$.
Retained widths are adjusted to preserve the same total removal
budget; $K=1.5N$ is rounded to 11. Smaller $K$ concentrates
pruning in fewer MLPs.

\subsection{Additional Accuracy Ablations}
\label{app:additional-accuracy-ablations}

These experiments follow the main evaluation protocol and reuse
previously evaluated controls. Avg.\ averages the displayed benchmark
means before rounding.

\paragraph{Calibration length.}
We vary the number of generated reasoning tokens used to collect
statistics on the same 24 AIME 2024 traces, retaining each complete
prompt. For \texttt{Qwen3-14B} at $N=7$, we recompute Block BI and feature
importance at each length while keeping $K=14$ and the parameter
budget fixed. Accuracy is stable beyond 1,024 tokens
(Table~\ref{tab:calibration-length}); we use 2,048 as the default.

\begin{table}[t]
\centering

\begin{adjustbox}{max width=\linewidth}
\begin{tabular}{@{}rrrrr@{}}
\toprule
Prefix & AIME25 & GPQA-D & HumanEval & Avg. \\
\midrule
512 & 52.2\std{3.8} & 52.9\std{1.3} & 87.8\std{2.7} & 64.3 \\
1,024 & 62.2\std{6.9} & 53.5\std{1.8} & 89.0\std{0.6} & 68.3 \\
2,048 & 64.4\std{5.1} & 49.7\std{2.9} & 90.2\std{1.6} & 68.1 \\
4,096 & 68.9\std{6.9} & 51.5\std{5.0} & 91.9\std{0.4} & 70.8 \\
\bottomrule
\end{tabular}
\end{adjustbox}
\caption{Calibration-length ablation on \texttt{Qwen3-14B} at $N=7$.
Prefix is the maximum number of generated reasoning tokens per trace;
each complete prompt is also included.
HumanEval reports pass@1 (\%); the other benchmarks report accuracy (\%).
The pruning budget and $K=14$ are fixed.}
\label{tab:calibration-length}
\end{table}

\begin{table}[H]
\centering
\begin{adjustbox}{max width=\linewidth}
\begin{tabular}{@{}lrrr@{}}
\toprule
Calibration source & AIME25 & GPQA-D & Avg. \\
\midrule
WikiText-2 & 0.0\std{0.0} & 20.0\std{2.3} & 10.0 \\
AIME prompts only & 22.2\std{6.9} & 38.2\std{4.7} & 30.2 \\
\rowcolor{oursbg}
AIME reasoning (Ours) & 64.4\std{5.1} & 49.7\std{2.9} & 57.1 \\
\bottomrule
\end{tabular}
\end{adjustbox}
\caption{Calibration-source ablation on Qwen3-14B at $N=7$.
Entries are accuracy (\%); Avg. averages the two benchmarks.
WikiText-2 and reasoning traces are token-matched;
prompt-only calibration uses fewer tokens.}
\label{tab:calibration-source}
\end{table}

\paragraph{Calibration source.}
We compare AIME 2024 reasoning traces, token-matched WikiText-2 text,
and the same 24 AIME prompts without generated continuations on
\texttt{Qwen3-14B} at $N=7$.
We recompute Block BI and feature importance for each source,
keeping the pruning budget and $K=14$ fixed.
Reasoning traces yield higher average accuracy than WikiText-2
or prompts alone (Table~\ref{tab:calibration-source}).


\paragraph{Cross-feature terms in feature selection.}
We compare feature-importance ranking with greedy selection based on
the squared MLP output change
(Equation~\ref{eq:analysis_mlp_output_change}).
Both use \texttt{Qwen3-14B} at $N=7$ with the same calibration traces,
BI-selected layers, and retained widths.
The squared output change includes cross-feature terms
$\mathbb{E}[\phi_i\phi_j]w_i^\top w_j$ ($i\ne j$),
which feature importance omits.
Both methods achieve 68.1\% average accuracy
(Table~\ref{tab:cross-term-accuracy}),
supporting feature importance as a lower-cost selection criterion
with comparable average accuracy.

\begin{table}[t]
\centering
\begin{adjustbox}{max width=\linewidth}
\begin{tabular}{@{}lrrrr@{}}
\toprule
Selection criterion & AIME25 & GPQA-D & HumanEval & Avg. \\
\midrule
Squared MLP output change
& 60.0\std{6.7} & 52.5\std{1.5} & 91.9\std{1.5} & 68.1 \\
\rowcolor{oursbg}
Feature importance (Ours)
& 64.4\std{5.1} & 49.7\std{2.9} & 90.2\std{1.6} & 68.1 \\

\bottomrule
\end{tabular}
\end{adjustbox}
\caption{Comparison of feature-selection criteria on Qwen3-14B at $N=7$.
HumanEval reports pass@1 (\%); the other benchmarks report accuracy (\%).
Both methods use the same removal budget.}
\label{tab:cross-term-accuracy}
\end{table}

Full-error calibration and selection take 13.2$\times$ as long as
diagonal ranking on the same hardware.

\section{Representation error and Attention Diagnostics Experimental Details}
\label{app:attention-surrogate-analysis}

\paragraph{Representation error.}
Figure~\ref{fig:layertrim_observations}(b),(d) compares dense and
pruned models on the same four held-out AIME25 traces.
Each trace is truncated to 2,048 total tokens, with 256 continuation
positions sampled for measurement. Final-hidden NMSE is
\[
    \frac{\sum_t\|h_t^{\mathrm{pruned}}-h_t^{\mathrm{dense}}\|_2^2}
         {\sum_t\|h_t^{\mathrm{dense}}\|_2^2}
\]
where $t$ ranges over measured positions across the traces.
In panel (b), we apply trimming to all 40 layers.

\paragraph{Attention measurements.}
Panel~\ref{fig:attention-surrogate-analysis}(a)--(b) measures
attention in the final two decoder layers (38 and 39) using three
dense-model AIME25 traces truncated to 32K total tokens.
Panel ~\ref{fig:attention-surrogate-analysis}(a) reports KL divergence from dense to pruned attention
distributions across pruning budgets, measured by the number of
removed parameters.
Both panels exclude attention to the first token and rescale the
remaining probabilities to sum to one.

\paragraph{Attention overlap.}
For each query token and attention head, we select the eight earlier
tokens that receive the highest attention weights from that query.
We select these tokens separately in the dense and pruned models.
Panel~\ref{fig:attention-surrogate-analysis} (b) shows the fraction of the dense model's top-eight tokens
that remain in the pruned model's top eight, grouped by how far
each token precedes the query.
Higher overlap means that pruning better preserves which earlier
tokens the model attends to.

\paragraph{Pruning budgets.}
Panel~\ref{fig:attention-surrogate-analysis}(b) uses approximately 1.5B removed parameters.
The component configurations and discrete-budget differences are
the same as the first group of Table~\ref{tab:ablations}, described
in Appendix~\ref{app:ablation-details}.

\paragraph{Feature importance and output change.}
Panel~\ref{fig:attention-surrogate-analysis}(c) compares summed
feature importance with the measured MLP output change across 70
combinations of layers and retained widths on four held-out AIME25 traces.
Each point corresponds to a feature set $R$ removed using the
calibration ranking.
For activation $\phi_j$ and output weight vector $w_j$, the horizontal
axis sums the importance of the removed features:
\[
    \sum_{j\in R}\mathbb{E}[\phi_j^2]\|w_j\|_2^2
\]
The vertical axis measures the squared MLP output change when these
features are removed together, using the same inputs:
\[
    \mathbb{E}\left\|\sum_{j\in R}w_j\phi_j\right\|_2^2
\]
Both quantities are averaged over the held-out inputs.
Feature importance ignores cross-feature terms, while the measured
output change includes them.
Their strong rank agreement shows that feature importance reliably
identifies which pruning choices cause larger output changes.

\section{Qualitative Analysis}
\label{sec:qual_analysis}

The following examples supplement Figure~\ref{fig:quant_analysis}
with excerpts from the evaluated reasoning and coding outputs.
They illustrate repetition, incoherent token fragments, code-format
failures, and arithmetic errors; they are not prevalence estimates.
In particular, lexical collapse need not satisfy the exact-token
repetition criterion in the main text, explaining LaCo's low detected
repetition despite frequent token-limit hits.

\onecolumn
\begingroup
\scriptsize
\begin{longtable}{@{}>{\raggedright\arraybackslash}p{0.12\textwidth}>{\raggedright\arraybackslash}p{\dimexpr0.88\textwidth-2\tabcolsep\relax}@{}}
\toprule
\multicolumn{2}{@{}p{\textwidth}@{}}{\textit{Question:} An electron is in the spin state $(3i,4)$. Find $\langle S_y\rangle$, given $\sigma_y=\left(\begin{smallmatrix}0&-i\\i&0\end{smallmatrix}\right)$. Options: A. $-12\hbar/25$; B. $12\hbar/25$; C. $-25\hbar/2$; D. $25\hbar/2$.} \\ \specialrule{0.5pt}{2pt}{0.2pt} \specialrule{0.5pt}{0.2pt}{2pt} 
Method & Original trace excerpts and qualitative observations \\ \midrule
\endfirsthead
\endhead
\endfoot
\endlastfoot
\rowcolor{densebg}\tracemethod{Dense}{2,880}{Terminated} & \begin{minipage}[t]{\linewidth}\raggedright
\tracegap{} 
Let psi = (3i/5, 4/5). Then sigma\_y = [[0, -i],[i, 0]].\par We compute sigma\_y |psi> = [[0, -i],[i, 0]] * (3i/5, 4/5)\textasciicircum{}T = (0*(3i/5) + -i*(4/5), i*(3i/5) + 0*(4/5)) = (-4i/5, i*(3i)/5). But note: i*(3i) = 3*(i\textasciicircum{}2) = -3. So second component becomes -3/5. So sigma\_y |psi> = (-4i/5, -3/5).
\tracegap{} 
Then expectation value of S\_y is (hbar/2)*(-24/25) = -12 hbar/25.
\tracegap{}
\traceobs{Normalization and the matrix product remain consistent through the answer.}
\end{minipage}
 \\
\midrule
\tracemethod{ShortGPT}{32,256}{Limit hit} & \begin{minipage}[t]{\linewidth}\raggedright
\tracegap{} 
The correct answer is letter letter option letter letter letter letter answer letter letter letter letter letter letter letter letter letter letter letter letter letter letter letter letter letter letter letter letter le
\tracegap{} 
negative polarity positive polarity negative polarity positive polarity negative polarity positive polarity negative polarity positive polarity negative polarity positive polarity 
\tracegap{}
\traceobs{Answer-format and polarity loops replace the scientific calculation.}
\end{minipage}
 \\
\midrule
\tracemethod{LaCo}{32,256}{Limit hit} & \begin{minipage}[t]{\linewidth}\raggedright
\begin{CJK*}{UTF8}{gbsn}
itness\_duperge hungerCreatorsgenerated Witness\_waycup pred Counter brushingonga cagescreatorBush.IM\tracesymbol{�}回Pot collectivelyWitnessm
\tracegap{} 
vaeonusdfa832.readlineLossUGCidalDigest HOLDERacdmumu recModifiedDate falsevale Peiadotent Circuit escapeurf Vitalgist Authority CHANNELembr.qualCas447 multiline Highestline tet RingsanaxIF 
\tracegap{}
\end{CJK*}
\traceobs{Fragmented tokens continue to the cap without an interpretable calculation.}
\end{minipage}
 \\
\midrule
\tracemethod{GRASP}{32,256}{Limit hit} & \begin{minipage}[t]{\linewidth}\raggedright
\tracegap{} 
I should answer answer. Now the letter letter is answer. It is " answer letter. The letter letter. So letter is " letter. But answer. So letter letter: letter. answer letter. So letter answer letter. Actually letter letter: answer. letter. 
\tracegap{} 
letter letter letter letter. letter letter letter. So letter. So letter. So letter. So letter. So letter. So l
\tracegap{}
\traceobs{Answer-letter language repeats without advancing the derivation.}
\end{minipage}
 \\
\midrule
\tracemethod{FLAP}{32,256}{Limit hit} & \begin{minipage}[t]{\linewidth}\raggedright
\tracegap{} 
(-3i)*(-4i) + (4)*(-3) = 12i\textasciicircum{}2 - 12 = 12*(-1) -12 = -24.
\tracegap{} 
lambda\textasciicircum{}2 - (i * (-i)) = lambda\textasciicircum{}2 - (-1) = lambda\textasciicircum{}2 + 1 = 0, so eigenvalues are \ensuremath{\pm} i.
\tracegap{} 
So the answer is -12hbar/25. That is option A.\par Therefore, answer A is the correct answer.\par I'll now provide answer A.\par Thus answer A: -12hbar/25.\par Thus answer: Option A.\par I'll now answer in the requested format: \textbackslash{}boxed\{A\} is the answer.\par But wait, but the computed result is -12/25. But the options are in terms of hbar. So the answer is -12hbar/25.\par I must check the reasoning:
\tracegap{}
\traceobs{The correct option appears, but reconsideration continues to the cap; the saved scorer marks this response correct.}
\end{minipage}
 \\
\midrule
\tracemethod{RESP}{32,256}{Limit hit} & \begin{minipage}[t]{\linewidth}\raggedright
\tracegap{} 
We are given an electron spin state (3i, 4). It is provided as spin state vector in spin space. The spin state is (3i, 4). But let's check: It is given as (3i, 4). Possibly the spin eigenstates are normalized? But not normalized. But we can treat as spin spin vector.
\tracegap{} 
\_spin spin spin spin spin spin spin spin spin\_spin spin spin spin spin spin spin spin spin spin spin spin spin spin spin spin spin spin spin spin spin spin spin\_spin spin spin spin spin spin spin spin spin spin
\traceobs{Relevant domain vocabulary gives way to a repeated token stream.}
\end{minipage}
 \\
\midrule
\tracemethod{CoCurve}{32,256}{Limit hit} & \begin{minipage}[t]{\linewidth}\raggedright
\tracegap{} 
I suspect that the question is: "An electron is in the spin state (3/i, 4)"? Actually "3i, 4" means maybe a spin state written as a combination of components? Possibly the spin state is represented as a state vector with coefficients?
\tracegap{} 
\ensuremath{\langle}psi| sigma\_y |psi\ensuremath{\rangle} = \ensuremath{\langle}psi| sigma\_y |psi\ensuremath{\rangle} = \ensuremath{\langle}psi| sigma\_y |psi\ensuremath{\rangle} = \ensuremath{\langle}psi| sigma\_y |psi\ensuremath{\rangle}
\tracegap{}
\traceobs{The expectation-value notation is repeated instead of evaluated.}
\end{minipage}
 \\
\midrule
\rowcolor{oursbg}\tracemethod{\textbf{LayerTrim (Ours)}}{1,253}{Terminated} & \begin{minipage}[t]{\linewidth}\raggedright
\tracegap{} 
But v\textasciicircum{}\ensuremath{\dagger} v = |3i|\textasciicircum{}2 + |4|\textasciicircum{}2 = (9) + (16) = 25.
\tracegap{} 
sigma\_y v = [0 * (3i) + (-i)* (4), (i)*(3i) + 0*(4)] = [-i*4, i*(3i)]. But i*(3i) = 3i\textasciicircum{}2 = 3*(-1) = -3. So sigma\_y v = (-4i, -3).
\tracegap{} 
Then v\textasciicircum{}\ensuremath{\dagger} sigma\_y v = (-3i)*(-4i) + (4)*(-3). Let's compute: (-3i)*(-4i) = 12i\textasciicircum{}2 = 12*(-1) = -12. And (4)*(-3) = -12. So total v\textasciicircum{}\ensuremath{\dagger} sigma\_y v = -12 -12 = -24. Then expectation value sigma\_y = -24/25.\par Then expectation value of spin S\_y = (1/2)*(-24/25)*hbar = -12*hbar/25. So the answer is option A: -12*hbar/25.
\tracegap{}
\traceobs{The input, normalization, and intermediate products remain linked to a completed answer.}
\end{minipage}
 \\*
\bottomrule
\caption{Qualitative analysis of example reasoning traces on GPQA-Diamond using \texttt{Phi-4-reasoning-plus} at $L_p/L_t=9/40$ and seed $0$. \tracegap{} denotes omitted text.}
\label{tab:reasoning_trace_gpqa}
\end{longtable}
\endgroup

\begingroup
\scriptsize
\begin{longtable}{@{}>{\raggedright\arraybackslash}p{0.12\textwidth}>{\raggedright\arraybackslash}p{\dimexpr0.88\textwidth-2\tabcolsep\relax}@{}}
\toprule
\multicolumn{2}{@{}p{\textwidth}@{}}{\textit{Question:} Complete \texttt{has\_close\_elements(numbers, threshold)}: return whether any two input elements differ by less than the threshold. The prompt example \texttt{([1.0, 2.0, 3.0], 0.5)} requires \texttt{False}.} \\  \specialrule{0.5pt}{2pt}{0.2pt} \specialrule{0.5pt}{0.2pt}{2pt} 
Method & Original trace excerpts and qualitative observations \\ \midrule
\endfirsthead
\endhead
\endfoot
\endlastfoot
\rowcolor{densebg}\tracemethod{Dense}{1,348}{Terminated} & \begin{minipage}[t]{\linewidth}\raggedright
\tracestage{Reasoning}
\tracegap{} 
So for each i in range(len(numbers)), and for each j in range(i+1, len(numbers)), compare numbers[i] and numbers[j].
\tracestage{Final response}
\tracegap{} 
\begin{Verbatim}[breaklines=true,breakanywhere=true,breaksymbolleft={},breakanywheresymbolpre={},breakanywheresymbolpost={},tabsize=4]
def has_close_elements(numbers: List[float], threshold: float) -> bool:
    for i in range(len(numbers)):
        for j in range(i + 1, len(numbers)):
            if abs(numbers[i] - numbers[j]) < threshold:
                return True
    return False
\end{Verbatim}
\tracegap{}
\traceobs{Distinct-index comparisons are preserved in the final function.}
\end{minipage}
 \\
\midrule
\tracemethod{ShortGPT}{337}{Terminated} & \begin{minipage}[t]{\linewidth}\raggedright
\tracestage{Reasoning}
\tracegap{} 
The function should check if any two numbers in the list are within the given threshold. The function should be written in a Python block with the proper indentation.
\tracestage{Final response}
\tracegap{} 
The function uses the `*` operator to check if any two numbers in the list are within the given threshold. The function returns `True` if any two numbers are within the given threshold, and `False` otherwise.\par The function is as described, and the code is correct.
\traceobs{The response claims to provide an implementation, but supplies no comparison logic.}
\end{minipage}
 \\
\midrule
\tracemethod{LaCo}{32,256}{Limit hit} & \begin{minipage}[t]{\linewidth}\raggedright
\tracegap{} 
\begin{CJK*}{UTF8}{gbsn}
imer Spo aggregateTimelineaslasca Jenner官 TC户攀筹progressbar共创的高度GD\tracesymbol{�}栊cket clam命运随手oidantom迎接系列产品 movement踩春节路上功课NPCasl人性化刹档埠勿官-
\tracegap{}
\end{CJK*}
\traceobs{These mixed-script fragments do not form an interpretable plan or function.}
\end{minipage}
 \\
\midrule
\tracemethod{GRASP}{32,256}{Limit hit} & \begin{minipage}[t]{\linewidth}\raggedright
from typing import List, float, 2.0, 3.0, 4.0, 0.0, 2.0, 0.0, 0.0, 0.0, 0.0, 0.0, 0.0, 0.0, 0.0, 0.0, 0.0, 0.0, 0.0, 0.0, 0.0, 0
\tracegap{} 
0.0: 0.0: 0.0: 0.0: 0.0: 0.0: 0.0: 0.0: 0.0: 0.0: 0.0: 0.0: 0.0: 0.0: 0.0: 0.0: 0.0: 0.0: 0.0: 
\traceobs{A malformed import leads into numeric repetition, rather than a function body.}
\end{minipage}
 \\
\midrule
\tracemethod{FLAP}{4,122}{Terminated} & \begin{minipage}[t]{\linewidth}\raggedright
\tracestage{Reasoning}
\tracegap{} 
So, I can sort the list first, then check each consecutive pair. If any pair has a difference less than the threshold, then the answer is True. Otherwise, it's False.
\tracestage{Final response}
\tracegap{} 
\begin{Verbatim}[breaklines=true,breakanywhere=true,breaksymbolleft={},breakanywheresymbolpre={},breakanywheresymbolpost={},tabsize=4]
def has_close_elements(numbers: List[float], threshold: float) -> bool:
    sorted_numbers = sorted(numbers)
    for i in range(len(sorted_numbers) - 1):
    if sorted_numbers[i+1] - sorted_numbers[i] < threshold:
    return True
    return False
\end{Verbatim}
\tracegap{}
\traceobs{The intended algorithm survives, but the final if and return statements are incorrectly indented.}
\end{minipage}
 \\
\midrule
\tracemethod{RESP}{3,176}{Terminated} & \begin{minipage}[t]{\linewidth}\raggedright
\tracestage{Reasoning}
\tracegap{} 
Therefore, if the list is sorted, then the minimum difference between any two elements is the minimum of the adjacent differences.
\tracegap{} 
\begin{Verbatim}[breaklines=true,breakanywhere=true,breaksymbolleft={},breakanywheresymbolpre={},breakanywheresymbolpost={},tabsize=4]
    if len(numbers) < 2:
        return False
    numbers_sorted = sorted(numbers)
\end{Verbatim}
\tracestage{Final response}
\tracegap{} 
\begin{Verbatim}[breaklines=true,breakanywhere=true,breaksymbolleft={},breakanywheresymbolpre={},breakanywheresymbolpost={},tabsize=4]
if len(numbers) < 2:
        return False
    numbers_sorted = sorted(numbers)
\end{Verbatim}
\tracegap{}
\traceobs{Correct relative indentation is lost when the full function is converted to a body-only completion.}
\end{minipage}
 \\
\midrule
\tracemethod{CoCurve}{115}{Terminated} & \begin{minipage}[t]{\linewidth}\raggedright
\tracestage{Reasoning (empty)}
<think> </think>
\tracestage{Final response}
\tracegap{} 
\begin{Verbatim}[breaklines=true,breakanywhere=true,breaksymbolleft={},breakanywheresymbolpre={},breakanywheresymbolpost={},tabsize=4]
    numbers = set(numbers)  # Remove duplicates to consider each number only once
    for num in numbers:
        if any(abs(num - x) < threshold for x in numbers):  # Check if any number is within the threshold of the given number
            return True
    return False
\end{Verbatim}
\tracegap{}
\traceobs{The code compiles, but compares each value with itself; it incorrectly accepts the prompt's first example.}
\end{minipage}
 \\
\midrule
\rowcolor{oursbg}\tracemethod{\textbf{LayerTrim (Ours)}}{2,699}{Terminated} & \begin{minipage}[t]{\linewidth}\raggedright
\tracestage{Reasoning}
\tracegap{} 
But the problem says "closer to each other than given threshold". So, if the difference is exactly equal to the threshold, then it's not considered. For example, if threshold is 0.5, and two elements are 0.5 apart, then it's not considered. So the condition is strictly less than.
\tracestage{Final response}
\tracegap{} 
\begin{Verbatim}[breaklines=true,breakanywhere=true,breaksymbolleft={},breakanywheresymbolpre={},breakanywheresymbolpost={},tabsize=4]
if len(numbers) < 2:
    return False
sorted_numbers = sorted(numbers)
for i in range(len(sorted_numbers) - 1):
    if sorted_numbers[i+1] - sorted_numbers[i] < threshold:
        return True
return False
\end{Verbatim}
\tracegap{}
\traceobs{The final body preserves strict comparison and consistent indentation in the original prompt context.}
\end{minipage}
 \\*
\bottomrule
\caption{Qualitative analysis of example reasoning traces on HumanEval using \texttt{Qwen3-14B} at $L_p/L_t=9/40$ and seed $0$. \tracegap{} denotes omitted text.}
\label{tab:reasoning_trace_humaneval} \\
\end{longtable}
\endgroup

\begingroup
\scriptsize
\begin{longtable}{@{}>{\raggedright\arraybackslash}p{0.12\textwidth}>{\raggedright\arraybackslash}p{\dimexpr0.88\textwidth-2\tabcolsep\relax}@{}}
\toprule
\multicolumn{2}{@{}p{\textwidth}@{}}{\textit{Question:} Let $N$ be the number of positive integer divisors of $17017^{17}$ that leave a remainder of $5$ when divided by $12$. Find the remainder when $N$ is divided by $1000$.} \\  \specialrule{0.5pt}{2pt}{0.2pt} \specialrule{0.5pt}{0.2pt}{2pt}  
Method & Original trace excerpts and qualitative observations \\ \midrule
\endfirsthead
\endhead
\endfoot
\endlastfoot
\rowcolor{densebg}\tracemethod{Dense}{3,907}{Terminated} & \begin{minipage}[t]{\linewidth}\raggedright
\tracegap{} 
7 * 11 = 77, 77 * 13 = 1001, 1001 * 17 = 17017.
\tracegap{} 
So indeed, the valid cases are: (a even, d odd, b even) and (a odd, d even, b odd). And each gives 9 possibilities for a, 9 for b, 9 for d. So total 9\textasciicircum{}3 + 9\textasciicircum{}3 = 2*729 = 1458. And then times 18 choices for c yields 1458*18 = 26244.\par Then remainder mod 1000 is 26244 mod 1000 = 244.
\tracegap{}
\traceobs{The two parity cases and independent exponent choices produce the correct count.}
\end{minipage}
 \\
\midrule
\tracemethod{ShortGPT}{32,256}{Limit hit} & \begin{minipage}[t]{\linewidth}\raggedright
\tracegap{} 
But wait, is that all? Let's verify product: 7 * 11 = 77, 77 *13= 100, 77*13= 77*10 + 77*3? Let me compute: 77 * 10=770, 77*3=231, so product= 101, so no. Let's compute properly: 7*11=77; 77*13= 77*10 +77*3 = 770 +231 = 1, 770+231= 1,101 indeed.
\tracegap{} 
And then 1,101 *17 equals? 1,101*17 = 1,101*10 = 11,101? Let me compute: 1,101*17 = 1,101*7= 7,207, oh sorry, let me compute properly: 1,101*17 = 1,101*10= 11,101, but then 1,101*17 = 1,101*(10+7) = 11,101 + 7,207 = 18,308, oh sorry, not exactly.
\tracegap{} 
divided by prime 17017 divided by prime 17017 divided by prime 17017 divided by prime 17017 divided by prime 17017 divided by
\traceobs{An attempted verification introduces new arithmetic errors and then stalls.}
\end{minipage}
 \\
\midrule
\tracemethod{LaCo}{32,256}{Limit hit} & \begin{minipage}[t]{\linewidth}\raggedright
reads acquaintance567eyJremball congressentyAAAIDEOanne chargerbatismars-PackINS sachDDminer drift compensation Painterpressu
\tracegap{} 
s Tape validenger Observable playEDGE downhill dashojufenpagedpaged acceptedplaetty tun Missspotify yields Delta Lower sweep monitor CommissionPARprotarkin monitored
\tracegap{}
\traceobs{No interpretable factorization or count is visible in these fragments.}
\end{minipage}
 \\
\midrule
\tracemethod{GRASP}{32,256}{Limit hit} & \begin{minipage}[t]{\linewidth}\raggedright
\tracegap{} 
17017 is exactly a product of 17, 19, and 13, but let's check: 17 times 19 and 13: 17: 19 and then 13? So the factors of 17017 is 13, 17, 19. But check: we see that 17 factor? Actually: 17017 is: product of 17, 19, 13? Let's check: 17017 is a product of 13, 17, 19. So the digits? Let's check: 13*17*19? Let's check: 17*19? Actually, 17 times 19 is 60 maybe.
\tracegap{} 
 yes, yes, yes, yes, yes, yes, yes, yes, yes, yes, yes, yes, yes, yes, yes, yes, yes, yes, yes, yes, yes, yes,
\traceobs{The proposed factors multiply to 4,199; verification collapses into repetition.}
\end{minipage}
 \\
\midrule
\tracemethod{FLAP}{11,066}{Terminated} & \begin{minipage}[t]{\linewidth}\raggedright
\tracegap{} 
I want to check one more possibility: Possibly the factorization of 17017 is different. Let's re-check: 17017 = 17 * 7 * 11 * 13, I should check: 17*7 = 119, 119*11 = 1309, 1309*13 = 17017. So that's correct.
\tracegap{} 
For any exponent from 0 to 17, there are 9 odd choices and 10 even choices.\par For the pattern (odd, even, even): - a odd: 9 choices, b even: 10 choices, c even: 10 choices. Total = 9 \ensuremath{\times} 10 \ensuremath{\times} 10 = 900.\par For the pattern (even, odd, odd): - a even: 10 choices, b odd: 9 choices, c odd: 9 choices. Total = 10 \ensuremath{\times} 9 \ensuremath{\times} 9 = 810.
\tracegap{} 
Thus the total number N is 1710 \ensuremath{\times} 18 = 30780.
\tracegap{}
\traceobs{Factorization is recovered, but a wrong parity count survives into the answer.}
\end{minipage}
 \\
\midrule
\tracemethod{RESP}{4,054}{Terminated} & \begin{minipage}[t]{\linewidth}\raggedright
\tracegap{} 
For exponent range 0 to 17, there are 18 possibilities for each exponent. Among these, the number of even numbers from 0 to 17: Even numbers: 0,2,4,...,16. That's 9 numbers. The number of odd numbers: 1,3,...,17. That's also 9. So for each exponent, there are 9 even and 9 odd possibilities.
\tracegap{} 
Now, remember that the exponent c for prime 13 can be any number from 0 to 17, which is 18 possibilities. So overall, the number of divisors of 17017\textasciicircum{}\{17\} that are \ensuremath{\equiv} 5 mod 12 is: 1458 * 18 = 1458 * 18. Let's compute that: 1458 * 18 = 1458 * (10 + 8) = 1458*10 + 1458*8 = 14580 + 11664 = 26244. Wait check: 1458 * 18 = 1458 * 2 * 9 = 2916 * 9 = 26244. Yes.
\tracegap{}
\traceobs{RESP also succeeds here, preserving both parity counts and the final multiplication.}
\end{minipage}
 \\
\midrule
\tracemethod{CoCurve}{10,382}{Terminated} & \begin{minipage}[t]{\linewidth}\raggedright
\tracegap{} 
17\textasciicircum{}1 \ensuremath{\equiv} 17 \ensuremath{\equiv} 17 \ensuremath{-} 12 = 4 (mod 12) \quad{}\quad{}For any exponent \ensuremath{\geq}1, note that 17\textasciicircum{}n always gives 4 (mod 12) since 4\textasciicircum{}k is congruent to 4 for k \ensuremath{\geq}1. Thus: \quad{}\quad{}if d = 0 \ensuremath{\rightarrow} 17\textasciicircum{}0 \ensuremath{\equiv} 1 (mod 12) \quad{}\quad{}if d \ensuremath{\geq} 1 \ensuremath{\rightarrow} 17\textasciicircum{}d \ensuremath{\equiv} 4 (mod 12)
\tracegap{} 
Actually, 1458 \ensuremath{-} 1000 = 1458? No, more clearly: 1458 \ensuremath{-} 1000 = 1458? I'll do simple subtraction: 1458 \ensuremath{-} 1000 = 558.
\tracegap{}
\traceobs{A wrong subtraction changes the residue cases; a second survives into the reported remainder.}
\end{minipage}
 \\
\midrule
\rowcolor{oursbg}\tracemethod{\textbf{LayerTrim (Ours)}}{5,234}{Terminated} & \begin{minipage}[t]{\linewidth}\raggedright
\tracegap{} 
So the number of ways for a, b, d is: (9 ways for a if even, 9 ways for b if even, 9 ways for d if odd) = 9*9*9 = 729, and (9 ways for a if odd, 9 ways for b if odd, 9 ways for d if even) = 729. So total is 729+729 = 1458. And then c (the exponent for 13) is free, 18 possibilities. So overall count is 18 * 1458. Wait, did I already incorporate c? Let's check:
\tracegap{} 
But earlier, I computed 729+729 = 1458, but that 1458 is only for a, b, d. Then multiply by 18 for c. So total count N = 18 * 1458 = 26244. Let's check: 1458 * 18 = 1458*10 + 1458*8 = 14580 + 11664 = 26244.
\tracegap{} 
So N = 26244. But then the problem asks: "Find the remainder when N is divided by 1000." So 26244 mod 1000 = 244. But wait, check: 26244 mod 1000 = 26244 - 26*1000 = 26244 - 26000 = 244.
\tracegap{}
\traceobs{The main count preserves both parity patterns and explicitly checks the independent factor of 18.}
\end{minipage}
 \\*
\bottomrule
\caption{Qualitative analysis of example reasoning traces on AIME 2026 using \texttt{Phi-4-reasoning-plus} at $L_p/L_t=7/40$ and seed $0$. \tracegap{} denotes omitted text.}
\label{tab:reasoning_trace_math} \\
\end{longtable}

\endgroup
\twocolumn

\section{Licenses and Terms of Use}
\label{app:licenses}
All models and datasets used in this study are publicly available
for research use. We follow their respective licenses and usage
conditions and use these resources solely for non-commercial research.

\paragraph{Models.}
Qwen3-14B~\cite{yang2025qwen3} and
Magistral-Small-2506~\cite{rastogi2025magistral}
are released under the Apache 2.0 License.
Phi-4-reasoning-plus~\cite{phi4reasoning} is released
under the MIT License.

\paragraph{Evaluation datasets.}
AIME2025~\cite{aime25} and AIME2026~\cite{aime26}
are distributed through the Math-AI benchmark releases
under the Apache 2.0 License.
The original AIME problems are subject to the copyright
of the Mathematical Association of America.
GPQA-Diamond~\cite{rein2023gpqa} is released under
the CC BY 4.0 License, with additional dataset access
and redistribution conditions.
HumanEval~\cite{humaneval} is distributed under
the MIT License, while MBPP+~\cite{mbpp} is
distributed through EvalPlus under the Apache 2.0 License.

\paragraph{Calibration data.}
We use AIME2024 problems for calibration, with reasoning
traces generated by the original dense models.
The original examination problems are subject to the
copyright and usage conditions of the Mathematical
Association of America. The generated reasoning traces
are used for calibration and diagnostic analysis.
The calibration-source ablation also uses WikiText-2, whose dataset
card lists CC BY-SA 3.0 and GFDL licenses.\footnote{\url{https://huggingface.co/datasets/Salesforce/wikitext}}

\paragraph{Software.}
Hugging Face Transformers is released under the
Apache 2.0 License, while PyTorch is distributed
under a BSD-style license. All software is used
in accordance with its respective licensing terms.

\end{document}